\documentclass[review]{elsarticle}

\usepackage{lscape}
\usepackage{subfigure}
\usepackage{booktabs}
\usepackage[dvipsnames]{xcolor}
\usepackage{tikz}
\usepackage{array}
\usepackage{lineno,hyperref}
\usepackage{makecell}
\usepackage{multirow}
\usepackage{amsmath,mathtools,amsfonts,amsthm,fixltx2e}
\modulolinenumbers[5]

\begin{document}

\begin{frontmatter}

\title{Self-Supervised Representation Learning for Online Handwriting Text Classification}

\author[address]{Pouya Mehralian}
\ead{pouya.mehralian@ut.ac.ir}
\author[address]{Bagher BabaAli\corref{corresponding}}
\ead{babaali@ut.ac.ir}
\author[address]{Ashena Gorgan Mohammadi}
\ead{ashena.mohammadi@ut.ac.ir}
\address[address]{University of Tehran, Tehran, Iran}

\cortext[corresponding]{Corresponding author}


\begin{abstract}

  Self-supervised learning offers an efficient way of extracting rich representations from various types of unlabeled data while avoiding the cost of annotating large-scale datasets. This is achievable by designing a pretext task to form pseudo labels with respect to the modality and domain of the data. Given the evolving applications of online handwritten texts, in this study, we propose the novel Part of Stroke Masking (POSM) as a pretext task for pretraining models to extract informative representations from the online handwriting of individuals in English and Chinese languages, along with two suggested pipelines for fine-tuning the pretrained models. To evaluate the quality of the extracted representations, we use both intrinsic and extrinsic evaluation methods. The pretrained models are fine-tuned to achieve state-of-the-art results in tasks such as writer identification, gender classification, and handedness classification, also highlighting the superiority of utilizing the pretrained models over the models trained from scratch.

\end{abstract}


\end{frontmatter}

\nolinenumbers

\section{Introduction}\label{intro}

Supervised learning approaches have brought massive success in different fields of deep learning, owing a considerable part of their achievements to the high-quality large labeled datasets; however, there are downsides to this approach. Firstly, producing such datasets for each new task is expensive and time-consuming; for instance, it is impossible to collect a large dataset for many rare diseases; hence medical applications are supervision-starved. Secondly, semantic supervision can become an information bottleneck in the long run since there is no one-to-one correspondence between the distinct concepts in the sensory world and the finite sequence of words to describe them.

To mitigate the cost of collecting a large annotated dataset and diminish the effects of this bottleneck, a newfound and rapidly developing idea -- namely self-supervision -- uses large-scale unlabeled datasets to reach a high-level semantic understanding of the data at hand. Self-supervision refers to defining a puzzle -- called a pretext task -- on an unlabeled dataset and forming the pseudo labels as the appropriate answers. From here on, we can engage this problem in a supervised manner and hope the model learns a high-quality representation of the data. Many pretext tasks have been proposed for different domains over recent years. A few of them include jigsaw puzzle~\cite{noroozi2017unsupervised}, inpainting~\cite{pathak2016context}, colorization~\cite{zhang2016colorful}, classifying corrupted images~\cite{jenni2018selfsupervised}, part erasing~\cite{SPARE} for spatial domain, masked language model~\cite{devlin2019bert} and next sentence prediction~\cite{devlin2019bert} for natural language processing, and contrastive predictive coding~\cite{Oord2018RepresentationLW} and autoregressive predictive coding~\cite{APC} for speech domain. However, to the best of our knowledge, the self-supervised representation learning approach has not been researched on online handwriting text data.

Online handwriting data is the term used for the recorded trajectory of the portable pen on a digital screen. The extensive use of touchscreen electronic devices and portable pens in recent years has led to novel handwriting-based systems. These systems span a variety of applications, including digital forensics, personal authentication, behavior and demography categorization, and medical screening~\cite{biometric2012,plamondon2018,ELYACOUBI2019}. Hence, developing reliable and accurate analytical tools is vital in the modern world. While offline handwritten data, which refers to the scanned image of an individual's handwriting, is the main focus of most bodies of research in the digital handwriting data domain, online handwriting is the basis of more authentic and veracious analytical tools in numerous applications.

Datasets of handwriting can be categorized in two main groups: handwritten texts which include some written statements in a specific language, and handwritten documents which include a collection of texts, figures, and tables written by different individuals. IAM-onDB~\cite{iamondb}, BIT CASIA~\cite{casia}, IRONOFF~\cite{IRONOFF}, and CASIA-OLHWDB~\cite{olhwdb} are among the most reputable datasets in online handwriting text; although IRONOFF is rather an older dataset and has not been used in recent studies. While IAM-onDB is one of the largest bodies of pure online English handwriting text containing only the on-surface movements, the other two datasets are more popular in the Chinese and mixed Chinese-English language studies that provide the in-air movements along with additional features in each data point. Note that the pen's location and time on the surface are included in all mentioned datasets, while pressure or in-air trajectories of the pen are only available in CASIA or CASIA-OLHWDB datasets. On the other hand, Kondate~\cite{kondate} (Japanese texts, figures, and formulae), IAM-onDo~\cite{iamondo} (English documents) and CASIA-onDo~\cite{casiaondo} (the most recent English and Chinese document database) are categorized in the other group of handwriting dataset, CASIA-onDo being the largest. In terms of applicability, while handwritten documents might gain interest in behavioral and medical assessments, text datasets are still more popular and better suits for digital forensics and personal authentication. As a result, we will focus on the text datasets in this study.

In the literature on document analysis, handwriting-based tasks can be either text-dependent or text-independent~\cite{connel2002}. Unlike text-dependent tasks (such as signature verification), text-independent tasks are inherently more challenging and popular. Taking full advantage of today's technology of online handwriting acquisition systems, along with text-independent tasks, aids in introducing novel analytical tools for the aforementioned wide range of use. Various approaches can provide these tools, including deep learning methods.

Considering that self-supervised learning has become a prominent field based on its proficiency in learning high-quality representation from unlabeled data over the recent years, the proposal of methods to employ this learning paradigm in online handwriting text is beneficial. Some studies have used this learning paradigm in relevant domains to online handwritten data. Bhunia et al. offer two cross-modal pretext tasks for sketch and handwritten data: vectorization and rasterization~\cite{bhunia2021vectorization}. Vectorization maps images to the representing vector coordinates, and rasterization maps the vector coordinates to the corresponding images. These pretext tasks require the availability of both offline and online versions of the handwriting data simultaneously, and it is not possible to pretrain a model with only online handwritten data since these pretext tasks demand both modalities. To the best of our knowledge, there is no pretext task defined for online handwriting text data.

These motivated us to explore self-supervision on online handwritten data and propose our pretext task, Part Of Stroke Masking (POSM). We have defined two block structures, Full-state and Aggregate-state, which determine how to pretrain the models and extract high-quality representation utilizing this pretext task. Then, the pretrained models are evaluated on both intrinsic and extrinsic approaches. For the intrinsic evaluation approach, we depict the pretrained models' understanding of the patterns of characters on a task we named ``Reconstruction''. As for the extrinsic evaluation, firstly, two pipelines for fine-tuning the pretrained models on downstream tasks, namely the Inclusive and Exclusive pipelines, are suggested. Afterward, The pretrained models are evaluated on three downstream tasks: text-independent writer identification, gender classification, and handedness classification. We have achieved state-of-the-art results on all these tasks and shown the proficiency of POSM and our proposed methods for extracting rich representation from online handwriting text.

The rest of this study is organized as follows: In Section~\ref{sec:background}, we will review some works performed on handwritten data for the three mentioned tasks. Section~\ref{sec:methods} will mainly focus on describing the core representation learning methodologies employed in this study. Sections~\ref{sec:pretraining} and~\ref{sec:fine-tuning} focus on the proposed pretraining and fine-tuning procedures. Section~\ref{sec:experiments} explains the data employed in the study, as well as the intrinsic and extrinsic evaluation procedures methodically, and compares the results with the existing studies. Then, the efficiency of the learned representations is examined by comparing the learned representations of the pretrained model with a model trained from scratch. Later, in Section~\ref{sec:discussion}, some challenges are discussed, and a perspective of future works is provided. Finally, Section~\ref{sec:conclusion} will wrap up the study.


\section{Related Works}\label{sec:background}


\subsection{Writer Identification}

Writer identification addresses the classification of individual authors of a handwriting piece. Similarity distance computing and stochastic nearest neighbor algorithms were among the first methods tested on features extracted according to character prototype distribution for this task~\cite{chan2007,tan2008}. Meanwhile, the Gaussian mixture model (GMM) was introduced to create a universal background model~\cite{Liwicki_CASIA,schlapbach2008}. Later, Li et al. offered the method of hierarchical shape primitives for text-independent writer identification~\cite{li2009}. In 2013, dynamic time wrapping and a support vector machine (SVM) were employed after extracting features from the point, stroke, and word levels~\cite{gargouri2013}. However, this was not the only study using SVMs. In more recent works, an ensemble of SVMs was used for writer identification, in which each writer was modeled by an SVM using an adaptive sparse representation framework~\cite{Venugopal2020} and GMM-based feature representations~\cite{venugopal2019}. In addition, Ray et al. used an SVM with an LSTM-based encoder to identify the writers given a compressed representation of the data~\cite{ray2020}.

In addition to the mentioned methods, the so-called subtractive clustering was developed to identify the writing style of each individual~\cite{singh2015}. Later and 2017, Venugopal et al. adapted a codebook-based vector of a local aggregate descriptor to encode the sequence of feature vectors extracted at sample points, enhanced in their later work~\cite{venugopal2017,venugopal2018}. They also offered sparse coding-based descriptors~\cite{venugopal2018sparse} and a modified sparse representation classification framework, in which they introduce some a-priori information~\cite{venugopal2021}. In another recent work, an i-vector-based framework was proposed for the purpose~\cite{BabaAli2021}. In this novel framework, the sequence of the feature vectors extracted from each handwriting sample is embedded into an i-vector to derive long-term sequence-level characteristics for each writer. The classification phase includes within-class covariance normalization and regularized linear discriminant analysis followed by an SVM classifier.

The above approaches suggest novelties in the online writer identification task. Yet, none of them employs the power of artificial neural networks. The first use case of neural networks in this field was in 2015 when a feed-forward neural network was applied to features extracted by a Beta-elliptic model~\cite{dhieb2015}. The results improved further by making the neural network deep~\cite{dhieb2016}. Meantime, an end-to-end convolutional neural network-based approach was proposed for the purpose, offering data augmentation strategies~\cite{Yang_CASIA}. In 2017, recurrent neural networks (RNNs) were utilized, including Bidirectional Long Short-Term Memory (BLSTM) units, on the raw online handwritten data, providing an end-to-end framework with no hand-crafted features~\cite{zhang2017}. Later and in 2020, Dhieb et al. proposed a biometric system for multi-lingual text-independent writer identification, which feeds the result of feature extraction and grouping to a deep neural network~\cite{dhieb2020}. In 2021, stacked sparse autoencoders, followed by a softmax classifier proved to be accurate when applied to a series of features extracted by Beta-elliptic models and fuzzy elementary perceptual codes~\cite{dhieb2021}.

\subsection{Demography Categorization}

For online demography categorization, there are only a few studies available. In 2007, a benchmark was provided for gender and handedness classification on IAM-OnDB~\cite{liwicki2007}. This work combines online features with features extracted from the offline representation of the online data. Then, SVM and GMM are tested for the classification phase. The data is split based on individual writers, where data of 50 male and 50 female writers is used as the training set and 25 writers of each group as the test set for gender classification. The results depict the superiority of GMM over SVM, with a 67.06\% classification accuracy. For the task of handedness classification, since left-handed individuals are inherently less in number than the right-handed individuals in the given dataset, the split is done after picking as many right-handed individuals as there are left-handed writers. The same pattern of evidence as in gender classification is visible in this case as well, by the accuracy of 84.66\% using a GMM. To the best of our knowledge, this study is the only study reporting handedness on online handwriting.

Later, in 2011 and with inspiration from the previous work and slight modifications in features and the GMM parameters, online and offline data classifiers were combined, increasing the accuracy in gender classification to 67.57\%~\cite{liwicki2011}. In 2014, a self-organizing map (SOM) was suggested to classify writers’ gender from capital letters~\cite{faundez2014}. The study reports an accuracy of 76.00\% on BIOSECURID online handwriting dataset. In 2016, another text-dependent gender classification protocol was introduced, which regards strokes as the structural units of handwriting and uses separate pen-up and pen-down sequences as input features~\cite{sesa2016}. It employs SOMs to create independent pen-up and pen-down codebooks for male and female writers. In the most recent work in 2020, the cloud of line distribution (COLD) and Hinge features are coupled with two SVM classifiers (one per feature) to pursue the task on Arabic handwriting samples~\cite{gattal2020}.


\section{Proposed Framework for Representation Learning}
Our aim is to introduce a self-supervised representation learning method in order to extract rich representations from unlabeled online handwritten data. This is achievable by defining a suitable pretext task and forming the pseudo labels for the unlabeled data. In this Section, first, Part Of Stroke Masking (POSM) pretext task is presented, and then two network structures are suggested to solve this task, with the intention of learning high-quality representations from online handwritten data.
\label{sec:methods}

\subsection{Part of Stroke Masking (POSM)}\label{sec:posm}

The main idea behind our pretext task is to break each strokeset (a sequence of data points corresponding to a paragraph of handwritten text) into several normalized windows, then make a number of copies from each window and randomly mask a consecutive block in each of them, and train a model to predict the masked blocks. As previously mentioned, each strokeset contains several strokes. However, the length of the strokes varies since some characters are longer than others. To take advantage of neural networks, the data presented to them must be of equal size. To overcome this obstacle, each strokeset is divided into a number of windows of the same size.

For now, imagine a sequence of consecutive coordinates from the dataset, which is going to be transformed into a single masked window. First, this sequence is normalized and scaled. Then, a fraction of $f_{mask}$ of the data points are masked randomly, with the intention to train a model to reconstruct the entire window, thus predicting the masked part given the masked window. The masked part of the window should be a continuous block of data points; otherwise, it is often the case that the neighboring data points of a masked data point are not masked themselves, and it would be a simple task for the network to predict the masked data point, on the grounds that the neighboring data points are highly correlated and their values are close to each other. In this study, we empirically found that masking $0.3$ of each window works best, so the $f_{mask}$ parameter is set to $0.3$.

To construct the desired masked windows from the dataset, some preprocessing steps are suggested:

\subsubsection{Shift}\label{sec:Shift}

\begin{figure}[h]
  \centering
  \includegraphics[width=0.9\linewidth]{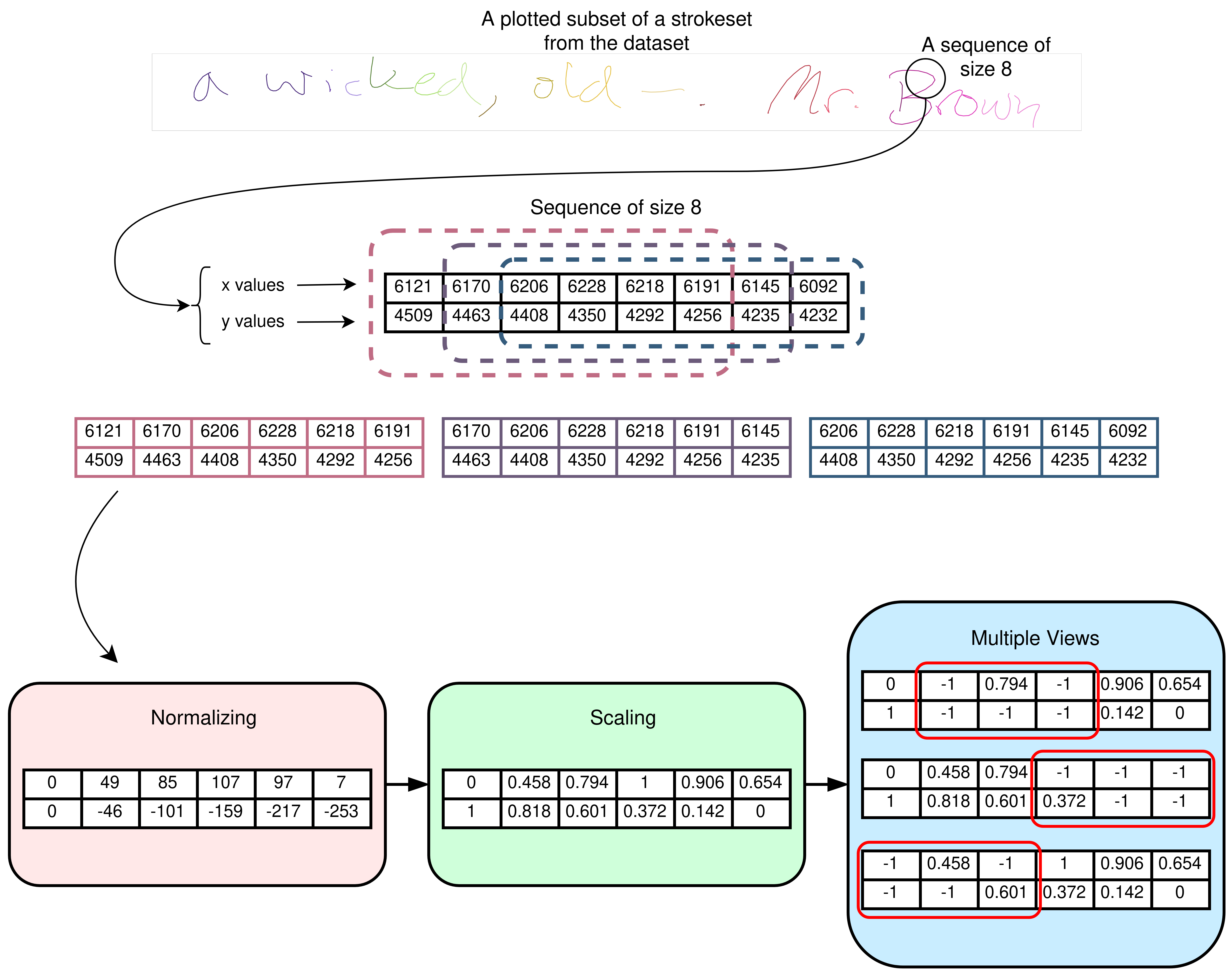}
  \caption{A numerical toy-example of the preprocessing procedure with $w_{size}=6$, $f_{mask}=0.5$ and $m_{views}=3$. Each window is normalized and scaled independent of the other windows. Afterward, $m_{views}$ copies of each window are made, and each of them is randomly masked.}
  \label{fig:windowing}
\end{figure}

Instead of simply dividing each strokeset into some sequences with no overlap, a pointer is put on the beginning of the strokeset, and then the first $w_{size}$ data points are taken as a window (i.e. $w_{size}$ is the size of the window). Next, the pointer is shifted $s_{posm}$ data points to the right, and the same is done ($s_{posm}$ is the shifting parameter). This continues until either the end of the strokeset is reached, or there would not be enough data points to construct a window if we shift. This preprocessing step is depicted in Figure~\ref{fig:windowing} which shows a toy example for the entire preprocessing procedure on a sequence of size 8, with $w_{size}$ set to 6 and $s_{posm}$ set to 1.

\subsubsection{Normalizing and Scaling}\label{sec:Normalizing_and_Scaling}

\begin{figure}[h]
  \centering
  \includegraphics[width=0.7\linewidth]{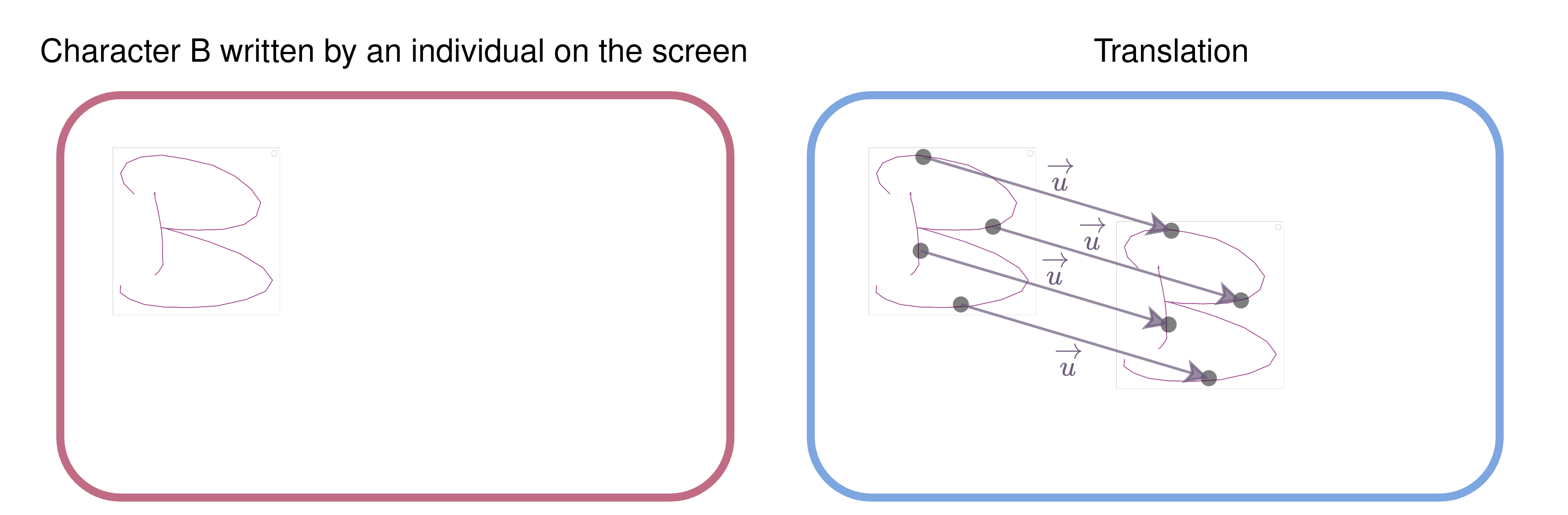}
  \caption{Moving character $\protect B$ on the screen by adding the vector $\protect \overrightarrow{u}$ to each coordinate of $\protect B$.}
  \label{fig:transformations}
\end{figure}

Considering the way the coordinates (x, y values) are being captured, it is apparent that the sequence representing a single character, like $B$, is directly related to where on the screen it is written. Let $c_1, c_2, \cdots, c_n$ be the longest sequence of coordinates, representing the character $B$ in the paragraphs of the dataset, and let $m$ be the total number of times the character $B$ has appeared in the dataset (either in different handwritings or different locations on the screen). Each of these $m$ instances of character $B$, with the length of $l$, can be represented by a vector in $\mathbb{R}^n \times \mathbb{R}^n$ by appending $n-l$, (0,0) coordinates to the end of its sequence.

It is possible to move about each instance of character $B$, with the original length of $l$ in the dataset, on the screen just by adding a vector like $\overrightarrow{u}  = (x_{u},y_{u}) \in \mathbb{R}^2$ to the first $l$ coordinates of its representing vector, without altering the pattern of handwriting (Figure~\ref{fig:transformations}). This would mean that a single character like $B$, in a specific handwriting, could be represented by infinitely many vectors in $\mathbb{R}^n \times \mathbb{R}^n$, depending on its location on the screen (in practice there are finitely many vectors since there are finitely many pixels on the screen; even so this number is very large). This, along with the limited number of paragraphs per individual in most handwriting text datasets, could cause the model difficulties in learning high-quality representations.



As a simple solution, the beginning of each window is moved to the zero point of the coordinate system. This is achieved in any window by subtracting the first data point of the window from the rest of the data points in that window. Afterward, each window is scaled using min-max scaling so that all the features have a value between 0 and 1.


\subsubsection{Multiple Views}\label{sec:multiple_views}

So far, for each window, a sequence of consecutive data points has been selected, normalized, and scaled. If each window were to be randomly masked once, the model would have looked at the parts that are not masked to predict the masked blocks. Depending on the $w_{size}$, the window might contain part of a character, a few characters, a single word, or a few words. Since each person's handwriting is unique and there are only a few short paragraphs from each subject, there are windows that rarely appear.

Let (S, H) be the sentence S of individual H’s handwriting, and imagine that S includes a word like ``Buzzword'', in which the character ``z'' is immediately followed by the ``w'' character. The number of words that contain ``zw'' in them is very small, and it is desirable that the model learns the pattern of how the character ``z'' might follow ``w''. Let $W=c_{1},c_{2}, \cdots,c_{w_{size}}$ be the raw window in the (S,H) pair, corresponding to the ``zw'' part in ``Buzzword'' (by raw, meaning that the window has been normalized and scaled, but has not been masked yet). If we were to make only one masked window from $W$, where $c_{i},c_{i+1},...,c_{j}$ are masked, the model would have looked at $c_{1},c_{2},...,c_{i-1}$ and $c_{j+1},c_{j+2},...,c_{w_{size}}$ to predict the masked block. It would be much better if the model was able to look at $W$, from multiple views. This is possible by making $m_{views}$ copies of $W$, and randomly choosing a block in each one to mask. This way, the model can learn the patterns in a deeper way, since it examines each raw window from multiple views, which results in a more robust model.

\begin{figure}[h]
  \centering
  \includegraphics[width=0.9\linewidth]{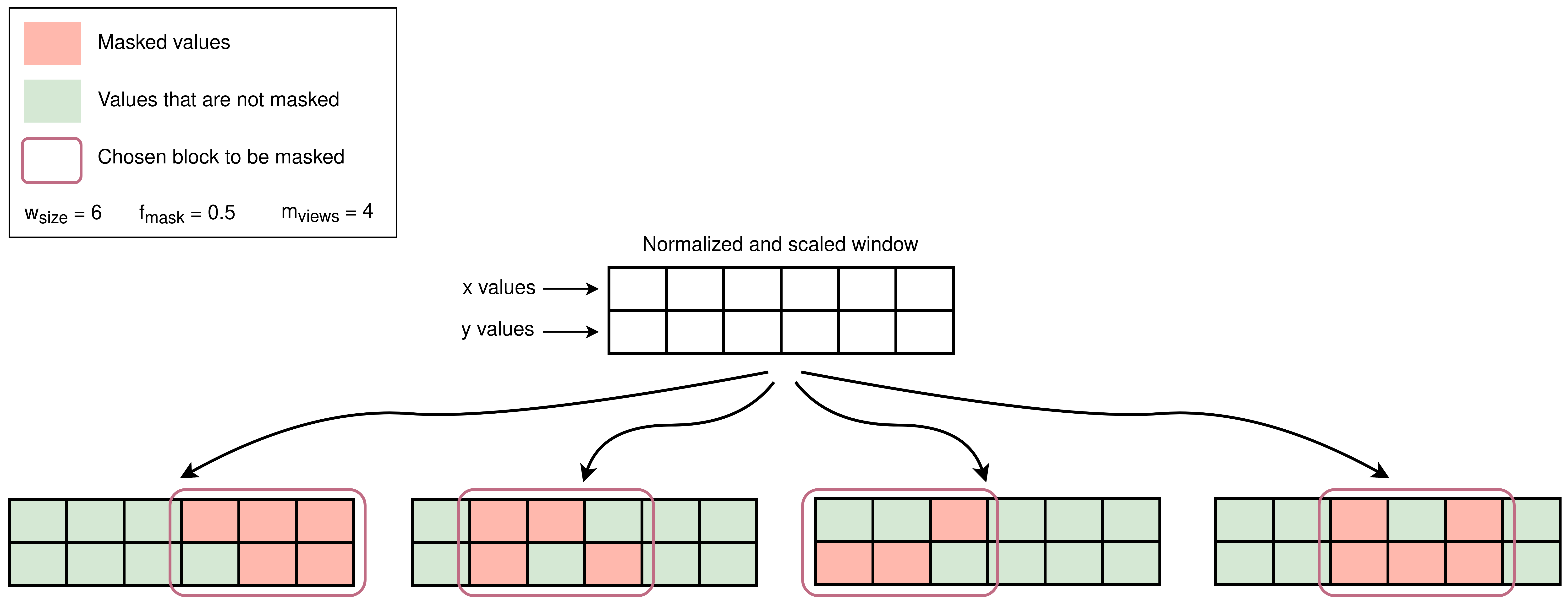}
  \caption{A toy-example of masking a normalized and scaled window with only x and y values as the input features, with $w_{size}$=6, $f_{mask}$=0.5, and $m_{views}$=4. Each of the features in each timestep of the chosen block is masked with the probabilty of 0.75.}
  \label{fig:masking}
\end{figure}

Now, to implement this idea, $m_{views}$ masked windows are formed from each raw window by randomly selecting $m_{views}$ blocks to be masked instead of one. Figure~\ref{fig:masking} demonstrates a toy example of the masking procedure. We call the $m_{views}$ parameter the multiple views parameter, the reason being that it will force the model to look at each chunk of the input sequence from multiple views. Since the values for each data point are between 0 and 1 after the normalization and scaling, any number outside of this interval can be chosen to represent a masked data point. In this study, -1 is chosen. A drawback to this approach is that -1 appears in the preprocessed dataset for pretraining but not in the downstream tasks. To diminish the effect of this issue, each of the features in each timestep of the chosen block is masked with the probability of 0.75; thus, avoiding a block containing only -1 values.

\subsubsection*{}

Intuitively, we can say that if the size of windows is set to be around the average length of a character, then the pretrained model focuses on learning the pattern of each character and how a single character might be written in different handwritings. Since the $s_{posm}$ parameter is set such that it would be smaller than $w_{size}$, this would result in forming plenty of windows in which there are some parts of two characters. Given enough data, the pretrained model will also learn the probability and how two characters might follow each other. In this study, the $w_{size}$ parameter is set to 32, which is approximately the size of an average character, the $s_{posm}$ parameter is set to 4, and the $m_{views}$ is set to 3.

\subsection{Proposed Network Structures For Pretraining}
\label{sec:structure}
In this subsection, two network structures are proposed, intended for pretraining the models on the POSM pretext task. These structures utilize LSTM layers as a building block that mitigates the vanishing gradient problem from which RNNs suffer. LSTMs are powerful in recognizing patterns in sequences of data and have been shown to be effective on online handwritten data in tasks like handwriting recognition~\cite{LSTM2,LSTM3} and end-to-end writer identification~\cite{LSTM1}.

\begin{figure}
  \centering
  \subfigure[]{
    \includegraphics[width=0.5030\linewidth]{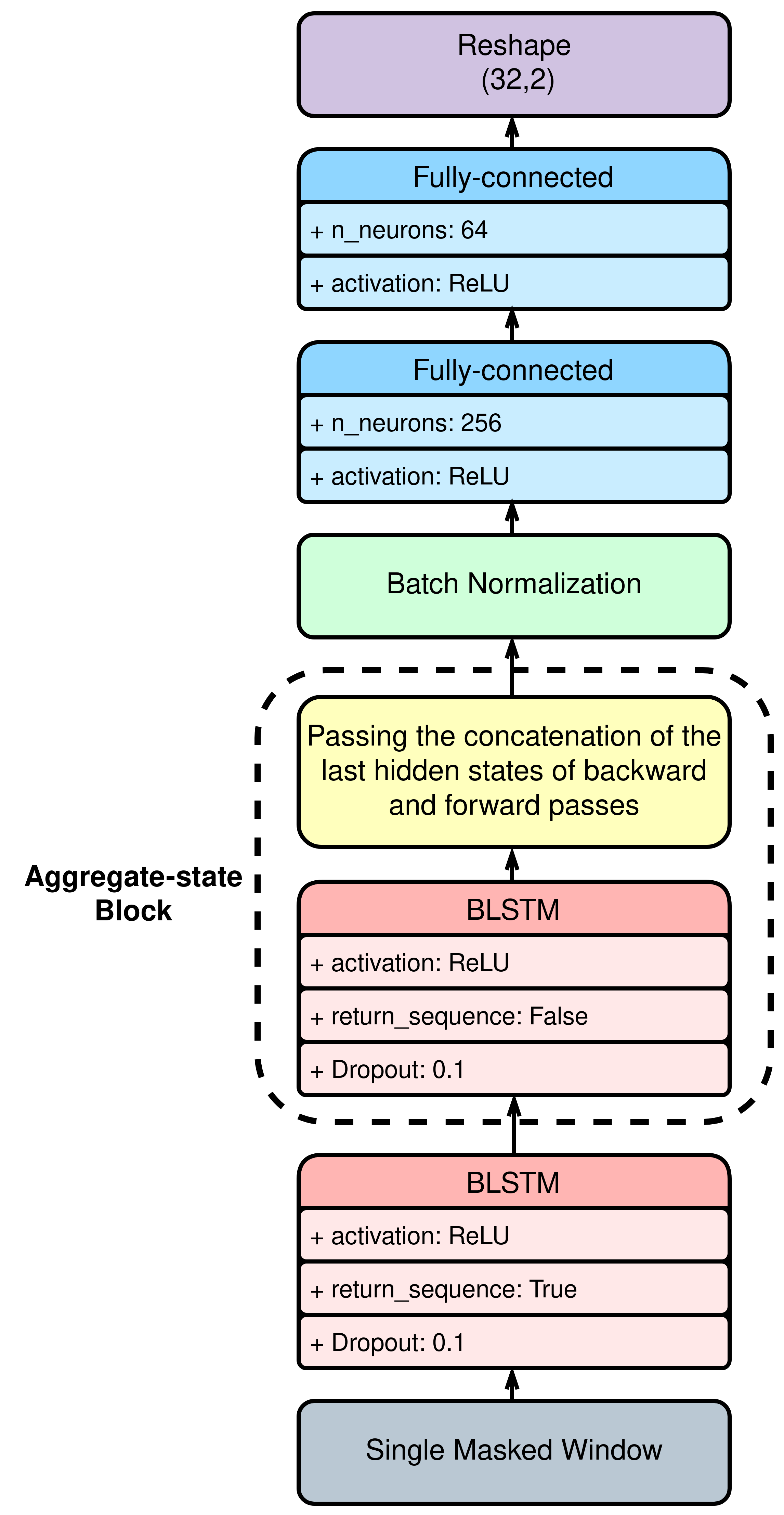}
  }
  \subfigure[]{
    \includegraphics[width=0.4390\linewidth]{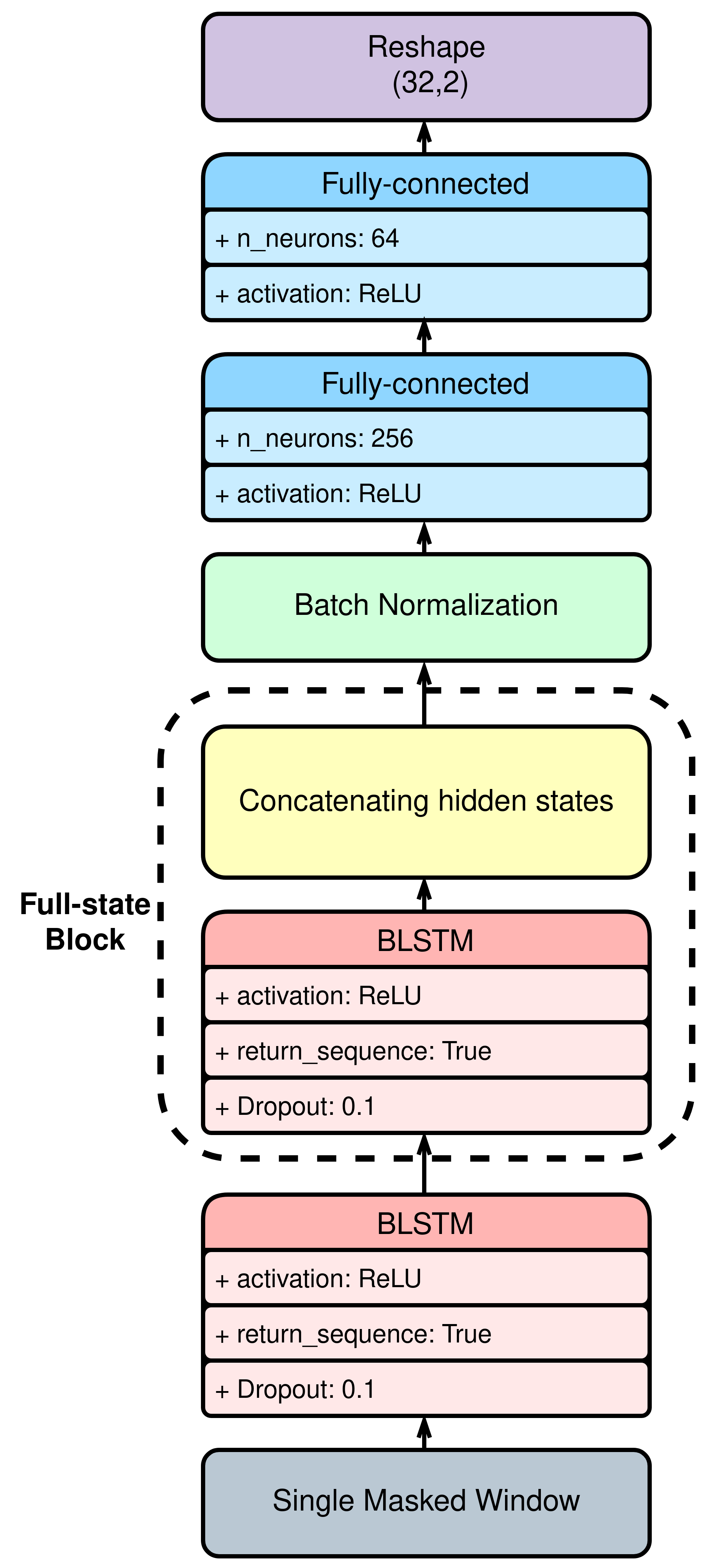}
  }
  \caption{An overview of the suggested structures for the pretrained models. (a) The pretrained model structure with Aggregate-state block (Aggregate-state structure). (b) The pretrained model structure with Full-state block (Full-state structure).}
  \label{fig:posmnet} 
\end{figure}

Since looking at the windows from both directions to predict the masked part is sensibly more effective than looking at either direction alone, it stands to reason that the model which looks at both directions will result in representations of superior quality. For this reason, we will use BLSTMs.  The two proposed network structures vary in how they process the output of the top BLSTM layer's hidden states to construct the input sequence's representation.
Depending on how the model produces the output representation of the top BLSTM layer, the structures are named. Regarding the BLSTM block, we either concatenate the last hidden state of the forward pass and backward pass of the top BLSTM layer as the representation of the input sequence (aggregate-state block) or concatenate all of the hidden states of the top BLSTM layer (full-state block). We will refer to these structures as ``Aggregate-state'' and ``Full-state'' respectively in that order. Figure~\ref{fig:posmnet} demonstrates the overview of the two structures.

We will show that a Full-state structure will result in a representation of superior quality compared to an Aggregate-state structure, on account of the fact that it provides more information across all parts of the window and avoids the information bottleneck if the size of the window is large (since in the aggregate-state block, the information of the entire window is compressed in one vector with a fixed size). Also, the subsequent feed-forward layers can learn to assign appropriate weights to each hidden state to predict the masked part. However, a Full-state structure is bigger in size compared to an Aggregate-state structure, given that the output representation of its BLSTM block is $w_{size}$ times longer. 

\begin{figure}[h]
  \centering
  \includegraphics[width=0.6\linewidth]{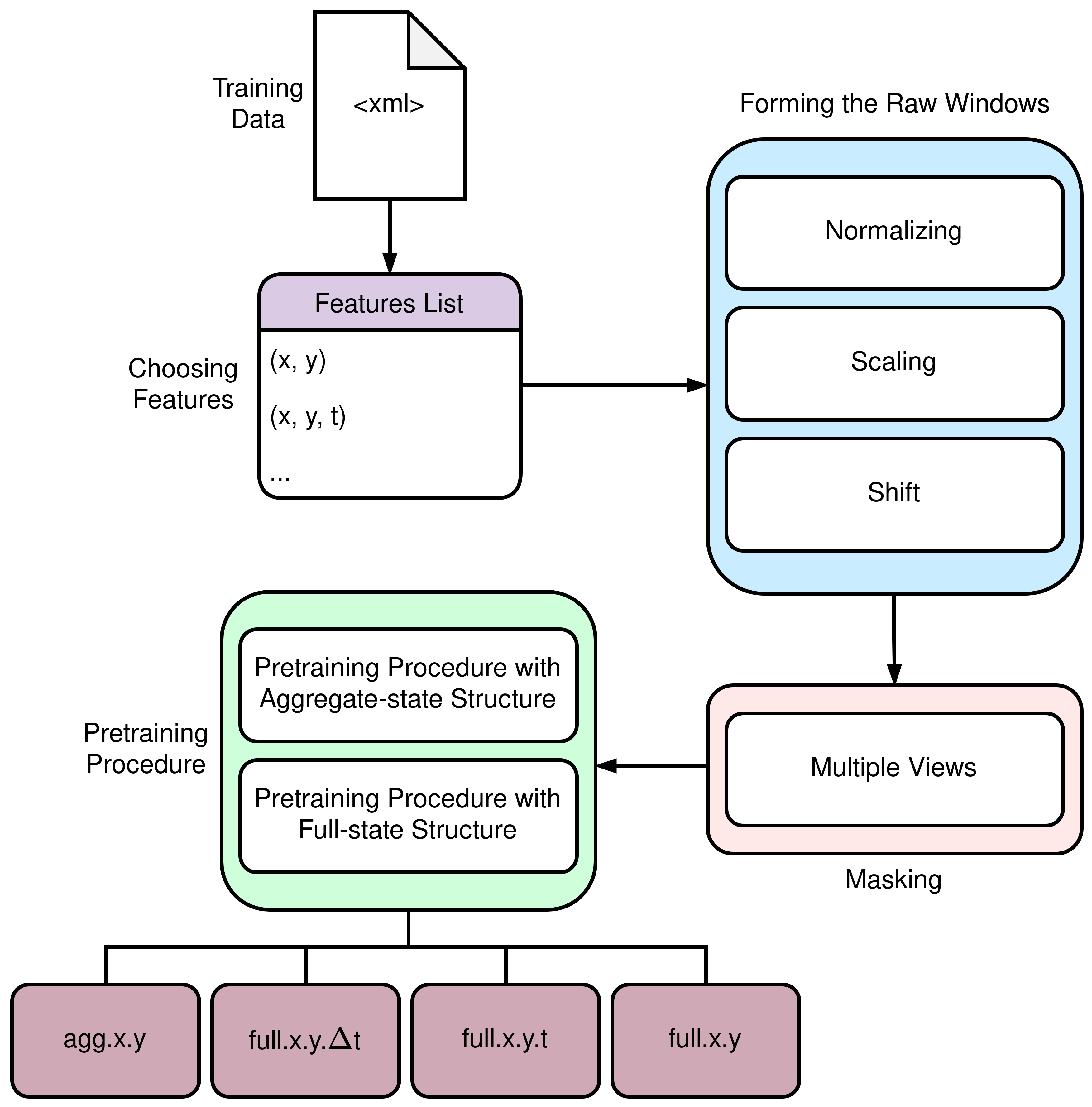}
  \caption{An overview of the pretraining pipeline.}
  \label{fig:pretrainPipeline}
\end{figure}


\section{Pretraining on POSM}
\label{sec:pretraining}

As mentioned earlier, each paragraph includes multiple data points through time. Specifically, each row of data includes a triplet $(x,y,t)$ where $x$ and $y$ indicate the horizontal and vertical position of the pen on the surface, and $t$ is the precise time of the positioning event. This information, along with the inherent sequential characteristic of the data, can be interpreted as either precise values or differential features. As a result, various choices are available for the selection of features for the pretraining purpose since any non-empty subset of $\{x, y, t, \Delta x, \Delta y, \Delta t\}$ can be picked. For each set of features, the preprocessing and forming of the windows would be the same except when we are dealing with $\Delta x$, $\Delta y$ or $\Delta t$, in which case, the beginning of the window should not be moved to the zero point of the coordinate system for these features. The experiments are performed on various feature sets, including $(x,y)$, $(x,y,t)$, $(x,y,\Delta t)$, $(\Delta x, \Delta y)$, and $(\Delta x, \Delta y, \Delta t)$. 


There are datasets that capture the in-air movements as well and provide a subset of additional features in each data point such as button status (usually a binary variable, being 0 for in-air movement and 1 for on-surface movement), the azimuth angle of the pen with respect to the tablet, altitude angle of the pen with respect to the tablet, and pressure applied by the pen. However, one major reason why we have chosen to only use the triplet of $(x, y, t)$ of only the on-surface movements from the raw data for the purpose of pretraining is the fact that this set is the biggest common subset of features of all the online handwriting text datasets available and is captured by all devices using a portable pen. The idea is that a model pretrained with feature sets such as $(x,y)$, $(x,y,t)$, $(x,y,\Delta t)$, $(\Delta x, \Delta y)$, and $(\Delta x, \Delta y, \Delta t)$ can be utilized for any online handwriting text dataset, but the same can not be said for the models pretrained with the mentioned additional features that are not provided in many datasets, such as IAM-onDB which offers only the triplet of $(x, y, t)$ of the on-surface movements.

After splitting the dataset to form the train set, the steps shown in the pretraining pipeline (Figure~\ref{fig:pretrainPipeline}) are followed, and the features for each model are selected. Then, depending on the choice of features, the masked windows are constructed for each of them. As Figure~\ref{fig:posmnet} shows, two main structures are used to pretrain various models on the POSM pretext task while modifying several parameters to reduce the mean squared error on the evaluation set. As it can be imagined, the number of distinct combinations of this process is significant due to the fact that there are multiple options for the set of features, as well as the parameters such as $f_{mask}$, $w_{size}$, $s_{posm}$ and $m_{views}$. Also, recall that there are two main suggested structures for the models, Aggregate-state structure and Full-state structure, each of which has many parameters as well. From here on, we focus on four of our pretrained models, which were able to extract representations of superior quality compared to others (Table~\ref{tab:pretrained}) and evaluate their performance on various tasks.

\begin{table}
  \caption{Network structures employed in the pretrained models. The naming provided in this table will be used in the rest of this study for ease of reference.}
  \label{tab:pretrained}       
    \begin{center}
    \footnotesize{
    \begin{tabular}{r|ccccc}
    \hline\noalign{\smallskip}
    Name & Block Type & Features & BLSTM Layers & FC Layers  \\
    \noalign{\smallskip}\hline\noalign{\smallskip}
    agg.x.y & Aggregate-state & $x, y$ & \parbox{2cm}{\centering 32*relu + 32*relu} & \parbox{3cm}{\centering BN + 256*relu + 64*relu + reshape(32,2)} \\ 
    \noalign{\smallskip}\hline\noalign{\smallskip}
    full.x.y.$\Delta t$ & Full-state & $x, y, \Delta t$ & \parbox{2cm}{\centering 32*relu + 32*relu + 32*relu} & \parbox{3cm}{\centering BN + 256*relu + 64*relu + reshape(32,2)}\\
    \noalign{\smallskip}\hline\noalign{\smallskip}
    full.x.y.t & Full-state & $x, y, t$ & \parbox{2cm}{\centering 32*relu + 32*relu + 32*relu} & \parbox{3cm}{\centering BN + 256*relu + 64*relu + reshape(32,2)}\\ 
    \noalign{\smallskip}\hline\noalign{\smallskip}
    full.x.y & Full-state & $x, y$ & \parbox{2cm}{\centering 32*relu + 32*relu + 32*relu} & \parbox{3cm}{\centering BN + 256*relu + 64*relu + reshape(32,2)}\\
    \noalign{\smallskip}\hline
  \end{tabular}
   }
\end{center}
  
\end{table}


\section{Fine-tuning the pretrained models on classification tasks}
\label{sec:fine-tuning}

\begin{figure}[h]
  \centering
  \includegraphics[width=0.88\linewidth]{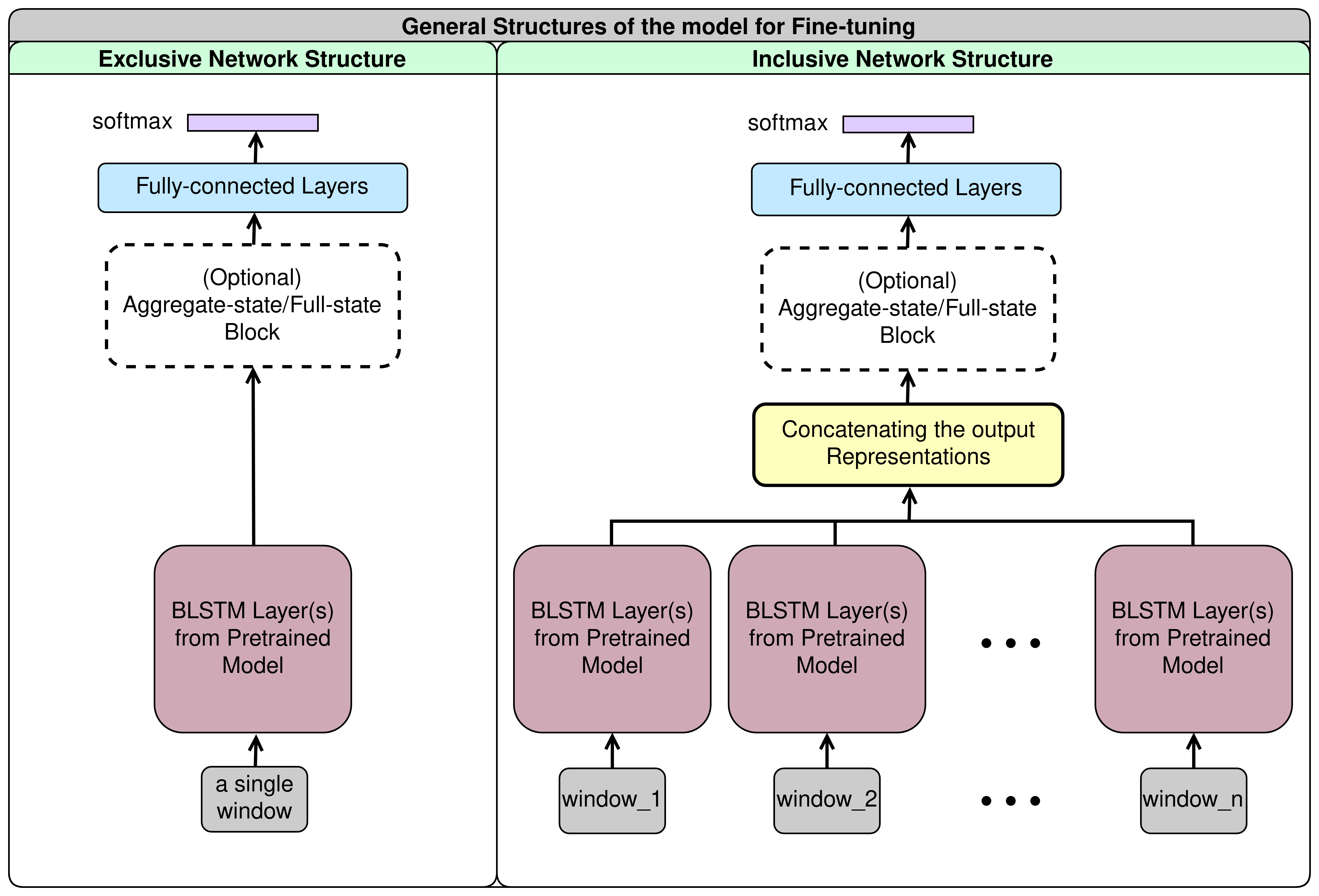}
  \caption{The general structures of the networks used in the downstream tasks. Note that the size of the softmax layer depends on the downstream task, where there would be one neuron per task class.}
  \label{fig:downstream}
\end{figure}

Imagine that we have been handed a training set for some supervised task on the online handwritten text and the training set includes several sequences of varying length, each with its specified label. Given that our pretrained models take inputs with a fixed size ($w_{size}$), the downstream models should do the same. This means that each sequence should be broken into chunks with the length of $w_{size}$ to feed them to the pretrained model. We propose two pipelines for fine-tuning the pretrained models, the Exclusive pipeline which utilizes the Exclusive network structure, and the Inclusive pipeline which utilizes the Inclusive network structure. Figure~\ref{fig:downstream} demonstrates these two network structures.

\begin{figure}[h]
  \centering
  \includegraphics[width=0.9\linewidth]{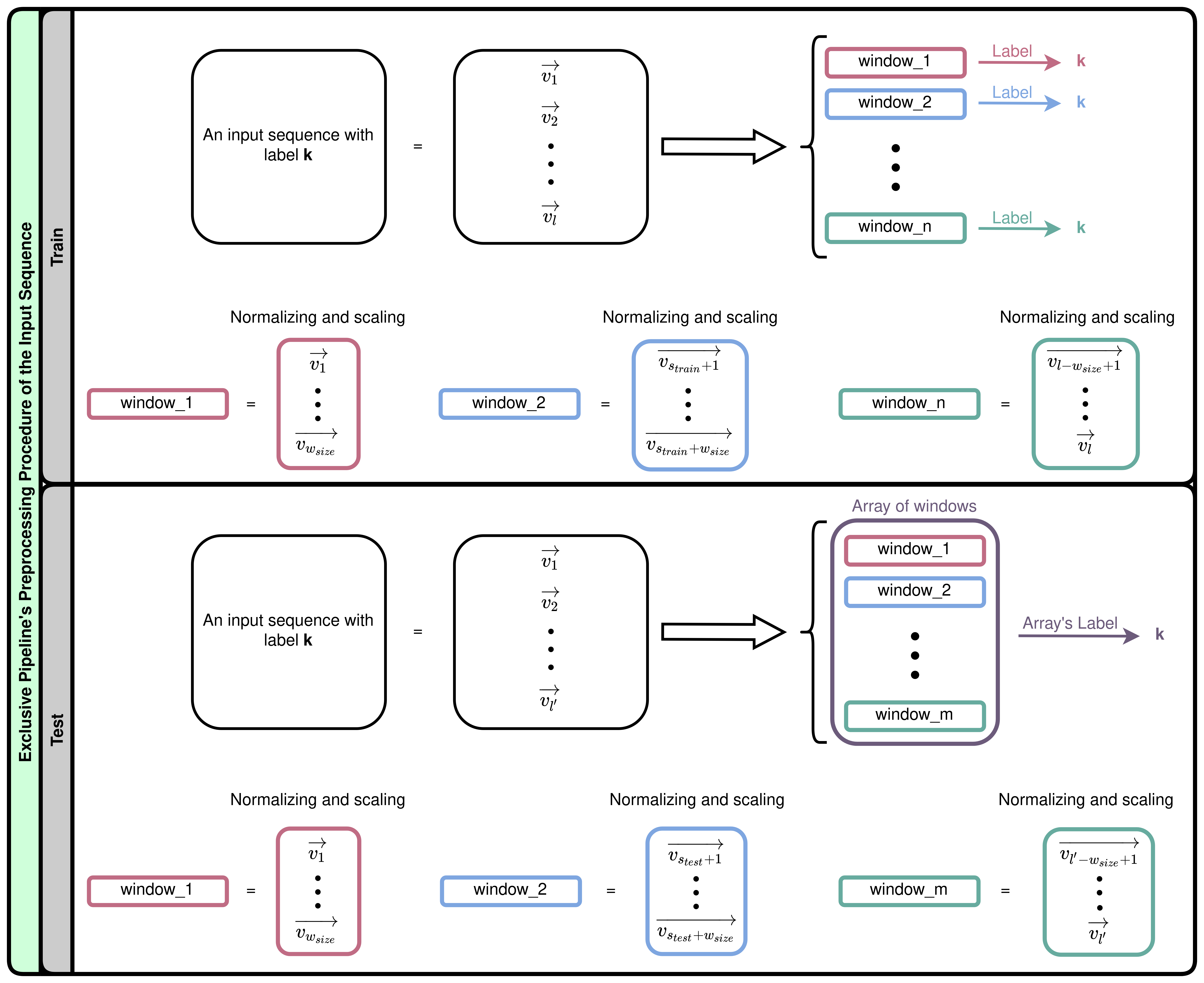}
  \caption{An overview of the Exclusive pipeline's preprocessing of the input sequences. $\protect \overrightarrow{v_{i}}$ indicates features vector corresponding to timestep $\protect i$.}
  \label{fig:PreExclusive}
\end{figure}

The Exclusive pipeline's preprocessing procedure of the input sequence starts from the beginning of each sequence in the train set and takes $w_{size}$ data points and performs the normalization and scaling described in Section~\ref{sec:Normalizing_and_Scaling} to form a single window. The label of this window is set as the label of the sequence. Then, this window is shifted $s_{train}$ data points to the right to form another labeled window as described in Section~\ref{sec:Shift}. The same steps are taken until either the end of the sequence is reached, or it is not possible to form another labeled window. By following this procedure, our desired preprocessed training set is formed. The top half of Figure~\ref{fig:PreExclusive} demonstrates the preprocessing procedure of the train set, in the Exclusive pipeline.

Regarding the preprocessing of the test set, the same steps are applied for each labeled sequence. However, instead of assigning a label to each window, for each sequence in the test set, an array is formed with all the windows corresponding to that sequence, and the label of the sequence is assigned to this array(bottom half of the Figure~\ref{fig:PreExclusive}). It should be noted that the shifting parameter of the test set ($s_{test}$) can be different from the shifting parameter of the train set ($s_{train}$), but the window size must be the same. The idea behind constructing the preprocessed test set in this manner is that our aim is to pass each of the windows belonging to an array corresponding to an input sequence in the original test set to the fine-tuned model and use soft voting on the results to predict the label of the array.


In the Exclusive pipeline, the Exclusive network is fine-tuned on the preprocessed train set. Then, for predicting the label for a given input sequence in the raw test set, each of the windows of its corresponding array in the preprocessed test set is passed to the trained model, and soft voting is performed on the results. In other words, for each input sequence, first, its corresponding array is constructed according to the preprocessing procedure of the Exclusive pipeline, then the sum of the outputs of the softmax layer of all the windows in the array is calculated. Finally, the predicted label will be determined by the calculated sum.

The model resulting from this fine-tuning pipeline is relatively small and thus is fast to train. However, there are some drawbacks, the reason being that the model only looks at a single window at a time. In each sequence that we want to classify, some windows are better indicators of the label of the sequence than the rest. Yet, the same weights are assigned to all of the windows in the summation stage. Also, many of the windows corresponding to an input sequence do not have sufficient information to classify it effectively. On these grounds, despite being light and fast, the Exclusive pipeline can be improved upon when it comes to performance.

\begin{figure}[h]
  \centering
  \includegraphics[width=0.9\linewidth]{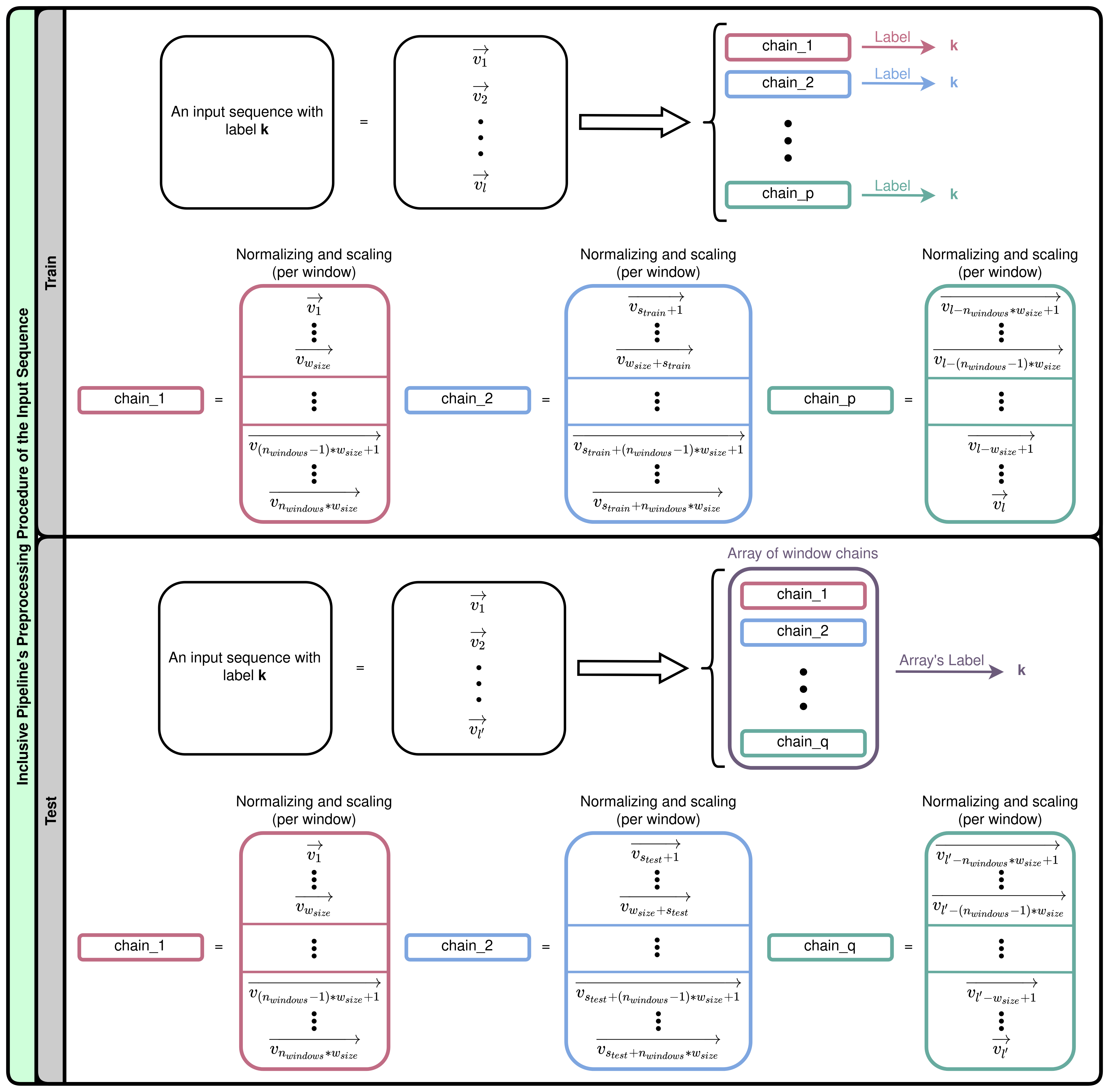}
  \caption{An overview of the Inclusive pipeline's preprocessing of the input sequences. $\protect \overrightarrow{v_{i}}$ indicates features vector corresponding to timestep $\protect i$. Each chain includes $\protect n_{windows}$ windows with no overlap ($\protect n_{windows} \times w_{size}$ timesteps). Each window in each chain is normalized and scaled separately.}
  \label{fig:PreInclusive}
\end{figure}

This motivated us to propose our second fine-tuning pipeline and the Inclusive network that considers multiple windows at a time, as shown on the right side of Figure~\ref{fig:downstream}. The preprocessing of the train set is different from the Exclusive pipeline. Here, for each labeled input sequence, instead of forming a single window at a time and shifting, the process will start from the beginning of each sequence and takes $n_{windows}\times w_{size}$ data points where $n_{windows}$ is the number of the windows that are going to be passed to the network at the same time. Then, $n_{windows}$ normalized and scaled windows with no overlap are formed from these data points. The label for this chain of windows is set as the label of its corresponding input sequence. Afterward, we will shift the beginning of each window $s_{train}$ data points and take another $n_{windows}\times w_{size}$ data points and do the same. The described steps are taken until either the end of the labeled sequence is reached, or it is not possible to form another chain of windows. The top half of Figure~\ref{fig:PreInclusive} demonstrates the preprocessing procedure of the train set, in the Inclusive pipeline.

In the Inclusive pipeline, a chain of windows is fed to the network. Afterward, the resulting representations of the input windows from the pretrained block are concatenated together, and the rest of the procedure is similar to the Exclusive network. In this pipeline, considering that the model takes several windows (15 to 20 windows in the downstream tasks of this study) instead of one, the number of characters in the input increases. As a result, such a network will decide how to engage the effect of multiple characters in the classification. In other words, it looks at all of the windows in the input and assigns a greater weight to those windows that are more effective in classifying the input sequence.

\begin{figure}[h]
  \centering
  \includegraphics[width=0.8\linewidth]{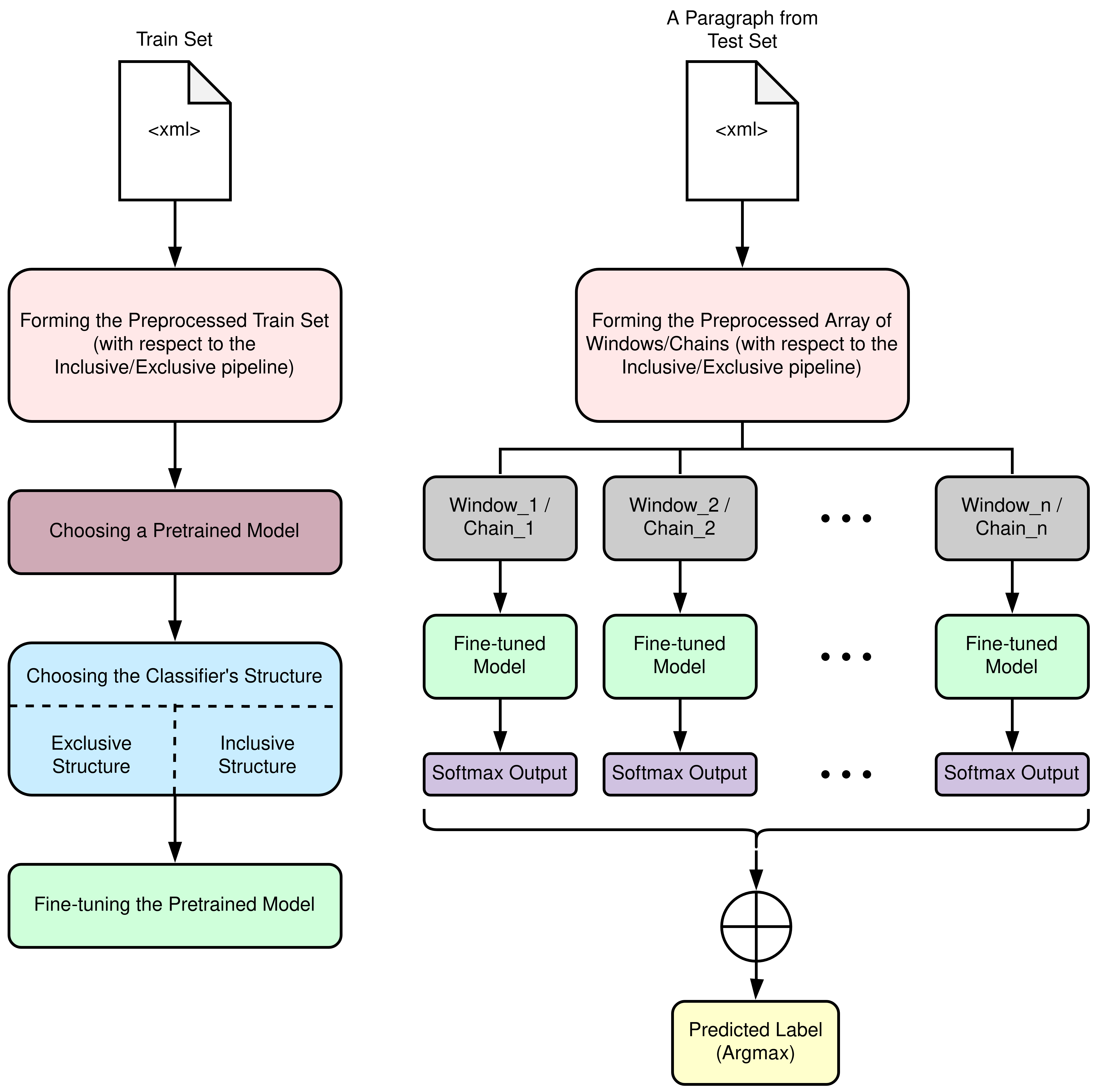}
  \caption{An overview of the proposed pipelines for fine-tuning the pretrained models and employing them in downstream tasks.}
  \label{fig:pipeline}
\end{figure}

The number of windows to be considered is one of the parameters which should be tuned to achieve the best performance. However, as this number evaluates the visibility range of the written text to the model, some assumptions can be made on how to define it. In these experiments, this number is set to 15 and 20, as these numbers provide a long-enough sequence of characters, spanning as long as 2 to 3 average-length words. Also, note that the representations are extracted from multiple windows in the course of the training/testing. This enables one to tune the pretrained layers for that specific task if needed. Figure~\ref{fig:pipeline} shows the entire pipeline of the downstream tasks.

We empirically found that setting the $s_{test}$ parameter to a small integer would improve the accuracy. This can be intuitively explained by comparing the shift size to the speed at which we move a magnifying glass with the diameter of $w_{size}$ throughout the paragraph. The slower we move, the more details we can capture; like focusing on the edges, the curves, the way the ending part of the strokes are dragged, and the pattern in which two characters follow each other. All these details will help us classify with more decisiveness.


\section{Experiments and Results}
\label{sec:experiments}

In this Section, the handwriting text datasets used in the study are briefly reviewed and then the efficiency of the pretrained models is examined on intrinsic and extrinsic evaluation approaches. Considering that our pretext task is unmasking the masked block of each window by reconstructing the entire masked window, one way to examine the pretrained models is to observe how they perform on the POSM pretext task on data they have not seen. The interpretation of the results on this task will function as our intrinsic evaluation of the pretrained models. As for the extrinsic evaluation, three downstream tasks have been chosen in which the pretrained models are fine-tuned on several classifiers to achieve state-of-the-art performances. Table~\ref{tab:networks} demonstrates the structure of these classifiers in detail. They either employ Inclusive or Exclusive structures with an optional Full-state or Aggregate-state block, followed by a BatchNormalization layer and a fully-connected block and using categorical/binary cross-entropy. Each classifier is named, depending on its structure.

\subsection{Data}
\label{sec:data}

As already mentioned, this study is taken out on online handwriting text data. In this study, IAM-OnDB~\cite{iamondb} is used to pretrain the models. This online handwriting database of English sentences is a baseline dataset for several handwriting tasks, such as handwriting recognition, writer identification, and gender and handedness classification. The database is gathered from 221 individuals with different originalities, genders, and writing types. There are eight handwritten paragraphs, each with about seven text lines, from each writer. Each text line is composed of multiple pen positions through time, grouped into different strokes, where each stroke is the data between each pen-down and pen-up event.

Given that the POSM pretext disregards the original labels of the data, using the entire IAM-onDB dataset for pretraining on the POSM pretext task should not cause any major issues when we want to evaluate the performance on a supervised task on the same dataset, the reason being that our pretrained models have not seen any actual labels. Additionally, there is a high chance that in the evaluation of the test set, the same windows would not be seen as the $s_{posm}$ parameter, and the shifting parameter used to form the windows of the test set might be different. However, to avoid any misgivings, we decided to train the pretrained models on only a subset of the dataset so that our pretrained models would not see even the raw test data of any downstream task in the stage of training on the POSM. To be more precise, the models are pretrained on 50 percent of the strokesets from each writer, meaning that four paragraphs from each subject have been selected such that they have no paragraphs or sentences in common with the test sets of any of the downstream tasks. 

In addition to IAM-OnDB, English datasets of the BIT CASIA~\cite{casia} database are employed in fine-tuning and writer identification task evaluation. Since BIT CASIA also contains handwriting texts in Chinese, both pretraining and fine-tuning were performed in this language. The collected samples in this dataset are captured by a Wacom Intuos2 tablet. The English dataset, written in the English language, includes 134 writers, and the Chinese dataset, written in Chinese, includes 187 writers. In both datasets, each writer has written a fixed sentence of about 50 words on one page and free-content sentences of about 50 words chosen freely by writers on two pages in the respective language. In each timestamp the following features are presented: $x$-coordinate, $y$-coordinate, time stamp, button status, azimuth, altitude, and pressure.

\subsection{Reconstruction}
\label{sec:reconstruction}
On account of the fact that the nature of the POSM pretext task is to predict the masked part of a window and considering that the $w_{size}$ of all the pretrained models (pretrained on IAM-onDB) in Table~\ref{tab:pretrained} are set to 32, which is approximately about the size of an average character, it should be examined if our pretrained models have learned the patterns of the characters and how they might be written in different handwritings. Also, since the $s_{posm}$ is set to four, many of the masked windows include parts of two characters; thus, we can observe if the pretrained models have learned how two characters might follow each other.

To accomplish this, some random windows on data that the pretrained models have not seen are formed. First, for each writer, a paragraph that was not used for either training or tuning the pretrained models' parameters is taken. Then, the preprocessing steps explained in Section~\ref{sec:posm} are followed in those paragraphs to construct a set of masked windows. Now, we examine and compare how the masked blocks of these windows are predicted. Figure~\ref{tab:table_of_figures} demonstrates the predictions of the pretrained models on some of these masked windows. As the figure shows, the model's performances are impressive. In all of the cases, the models successfully reconstructed the entire windows by recognizing the patterns of the masked blocks.

\newcommand{\addpic}[1]{\includegraphics[width=8em]{#1}}
\newcolumntype{M}[1]{>{\centering\arraybackslash}m{#1}}

\begin{figure}
    \centering
    \footnotesize{
    \noindent\makebox[\textwidth]{%
    \begin{tabular}{cM{7.2em}M{7.2em}M{7.2em}M{7.2em}M{7.2em}}
        \toprule
        Model. & a & b & c & d & e \\
        \midrule
        full.x.y & \addpic{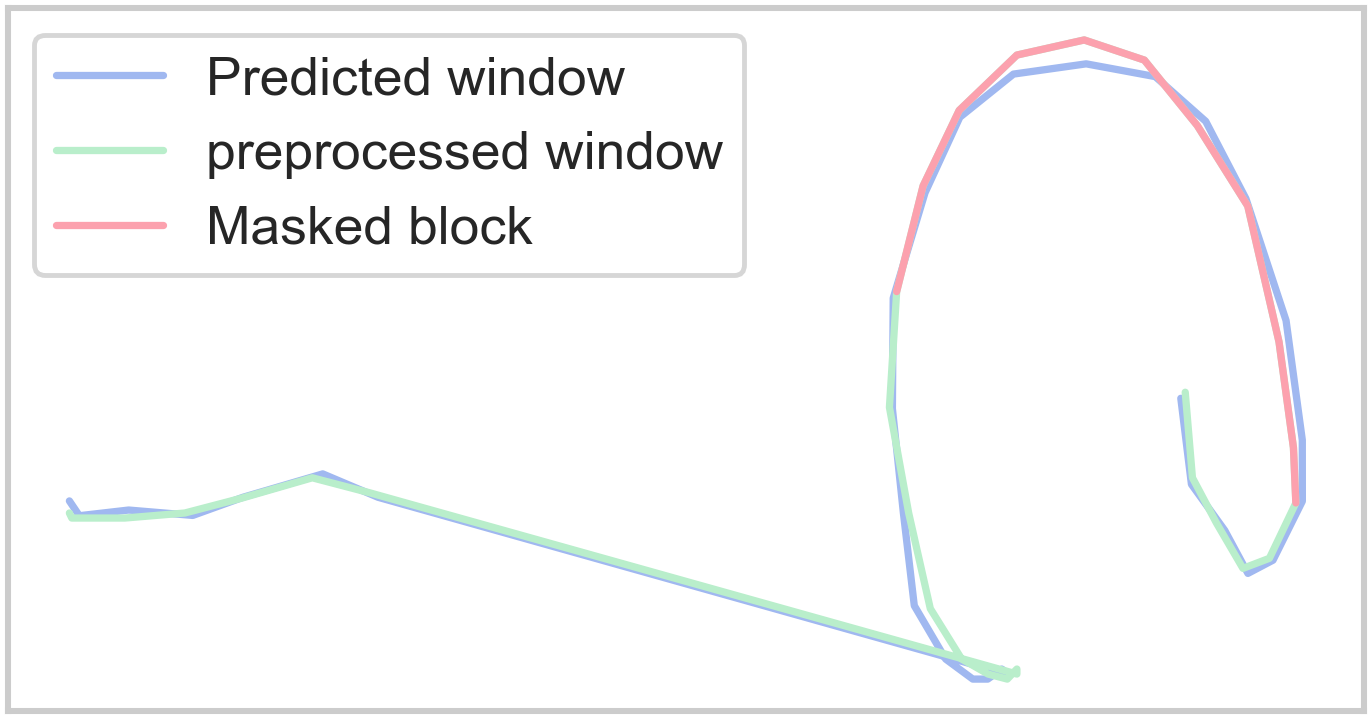} & \addpic{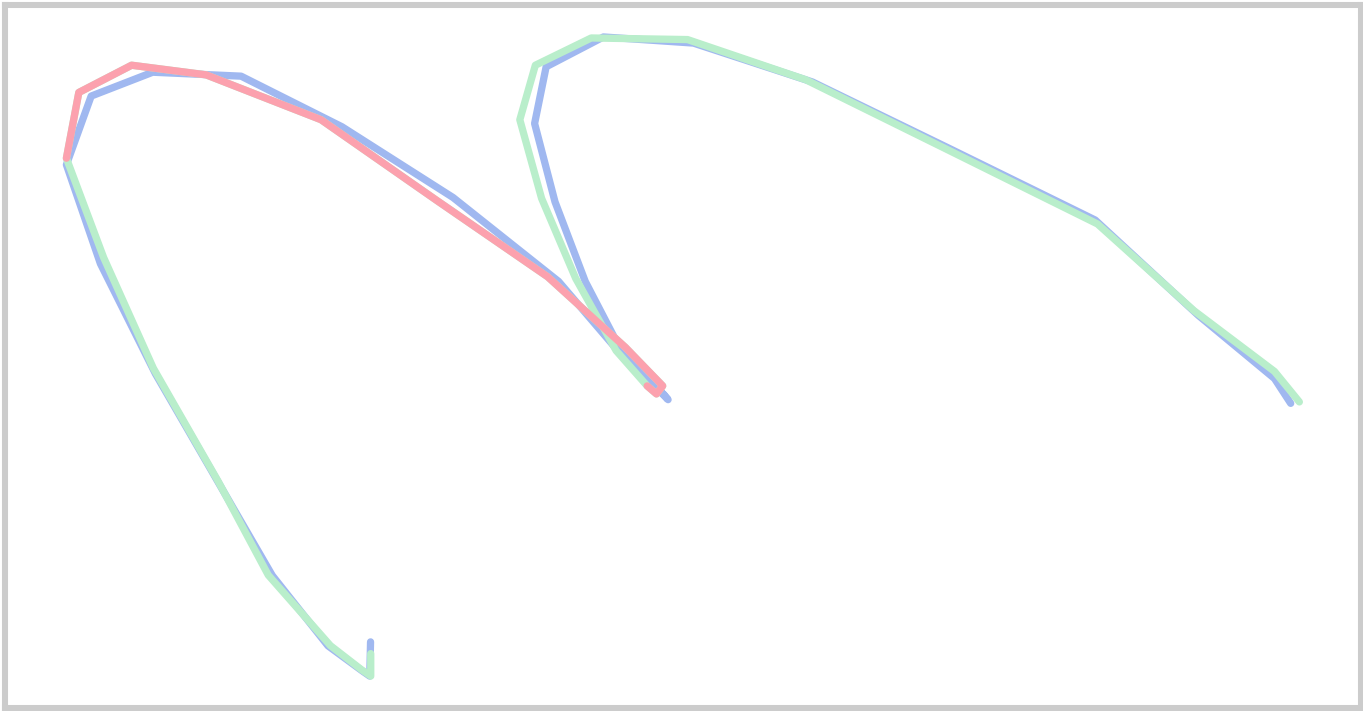} & \addpic{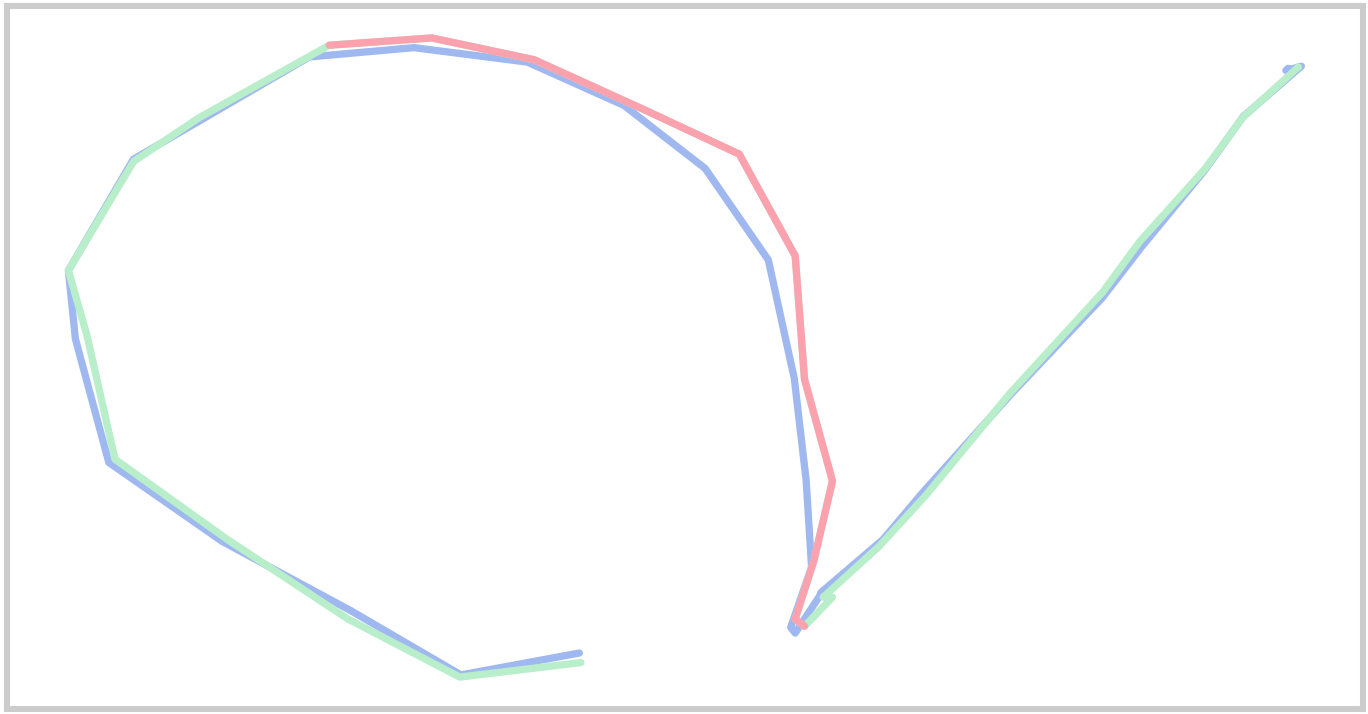} & \addpic{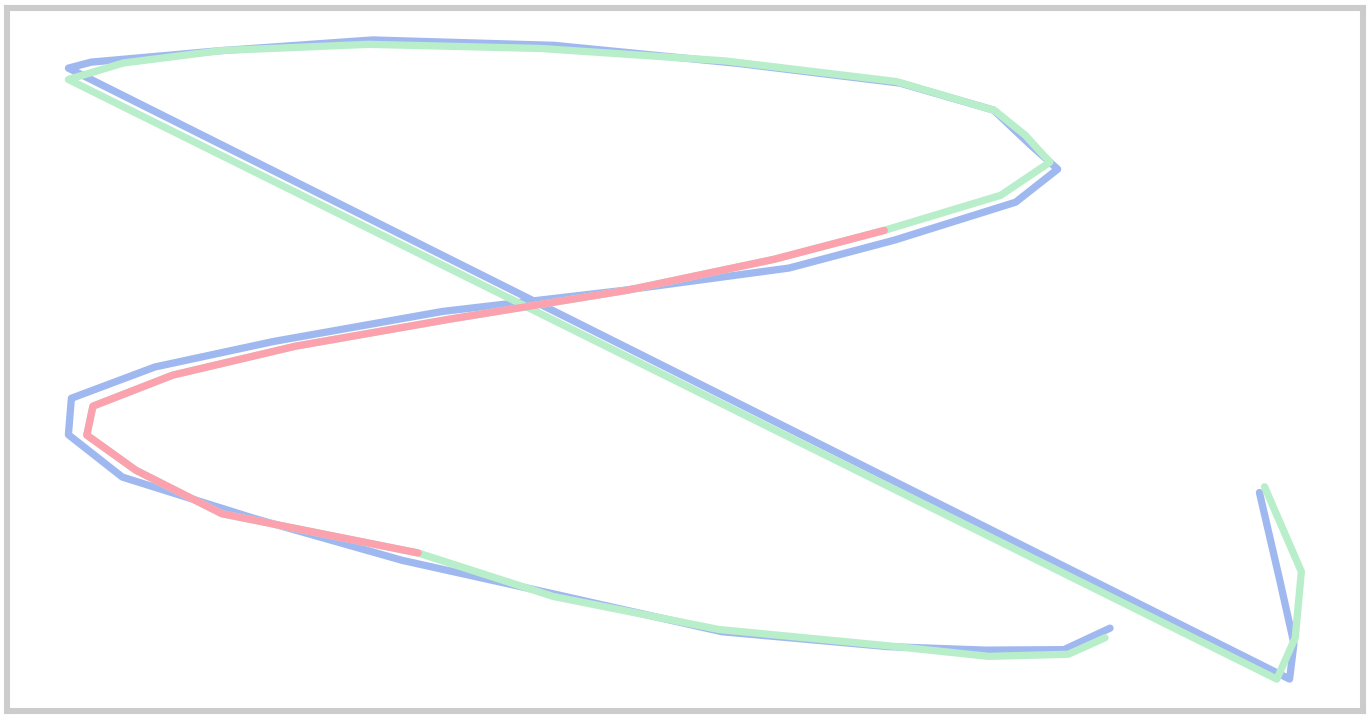} & \addpic{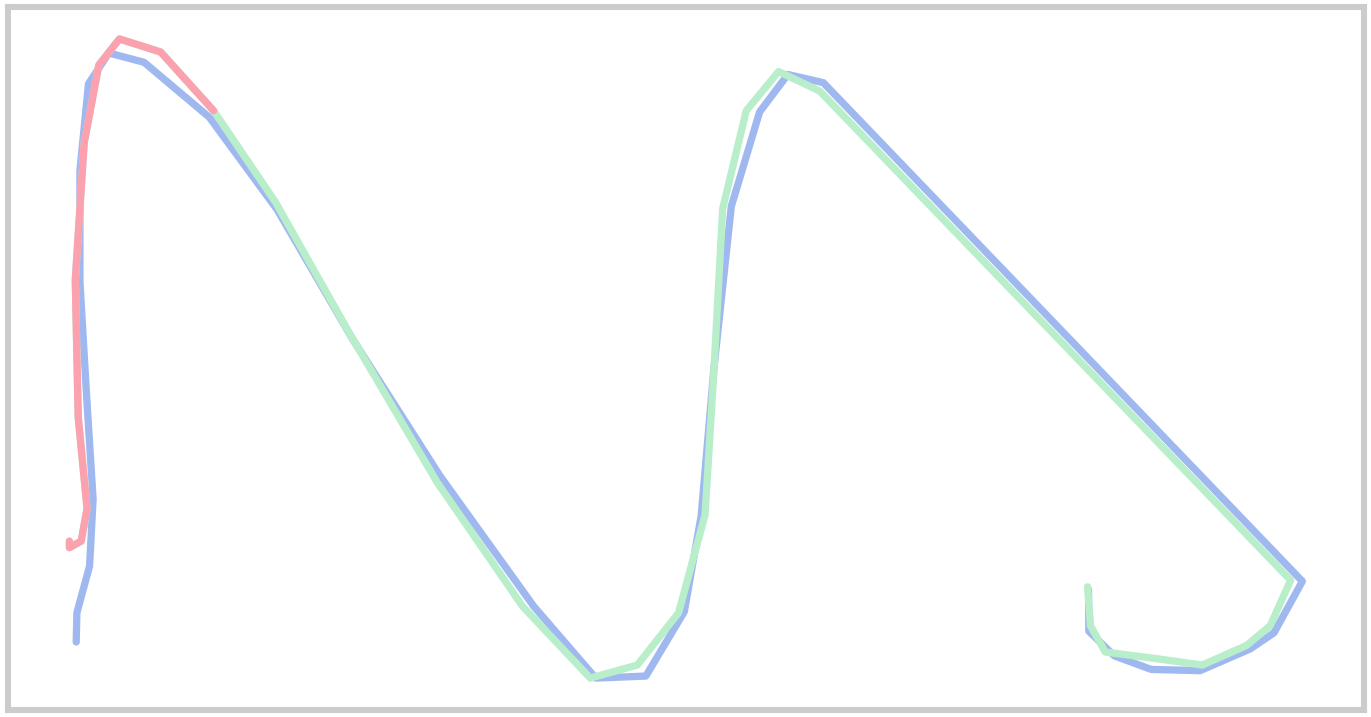}\\
        agg.x.y & \addpic{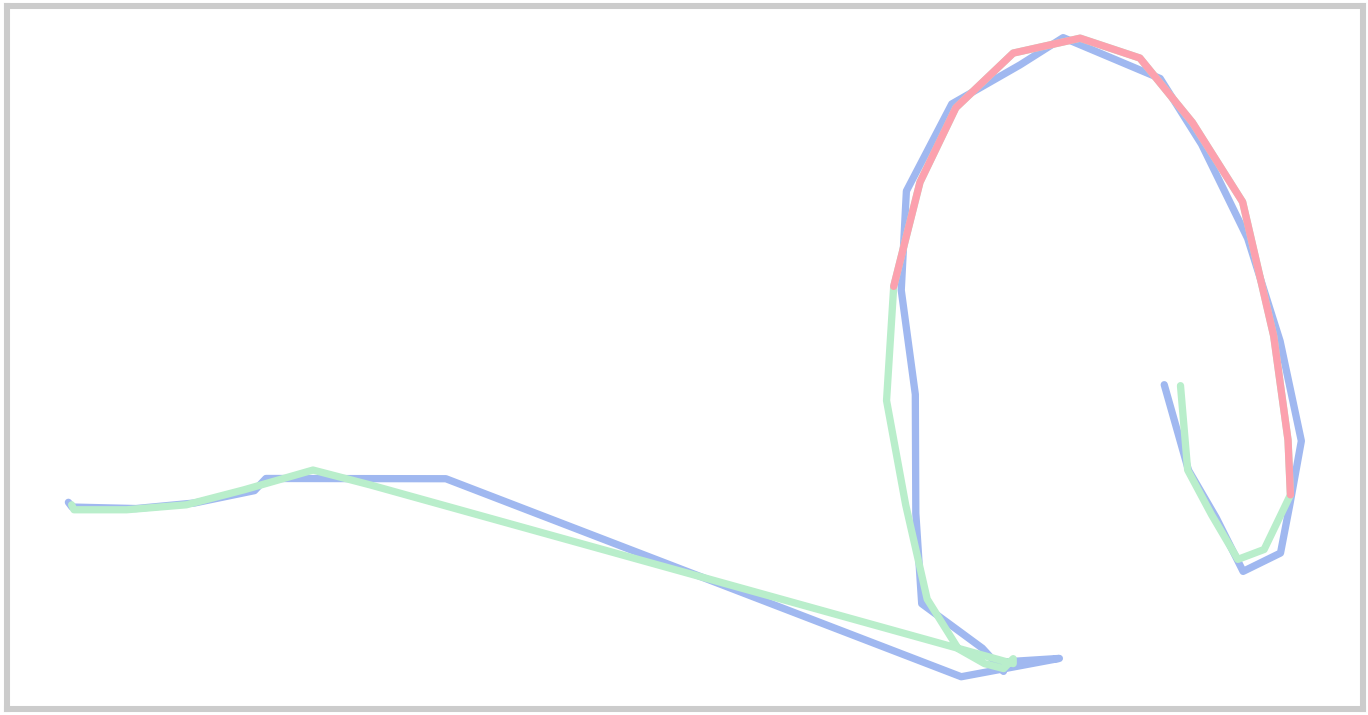} & \addpic{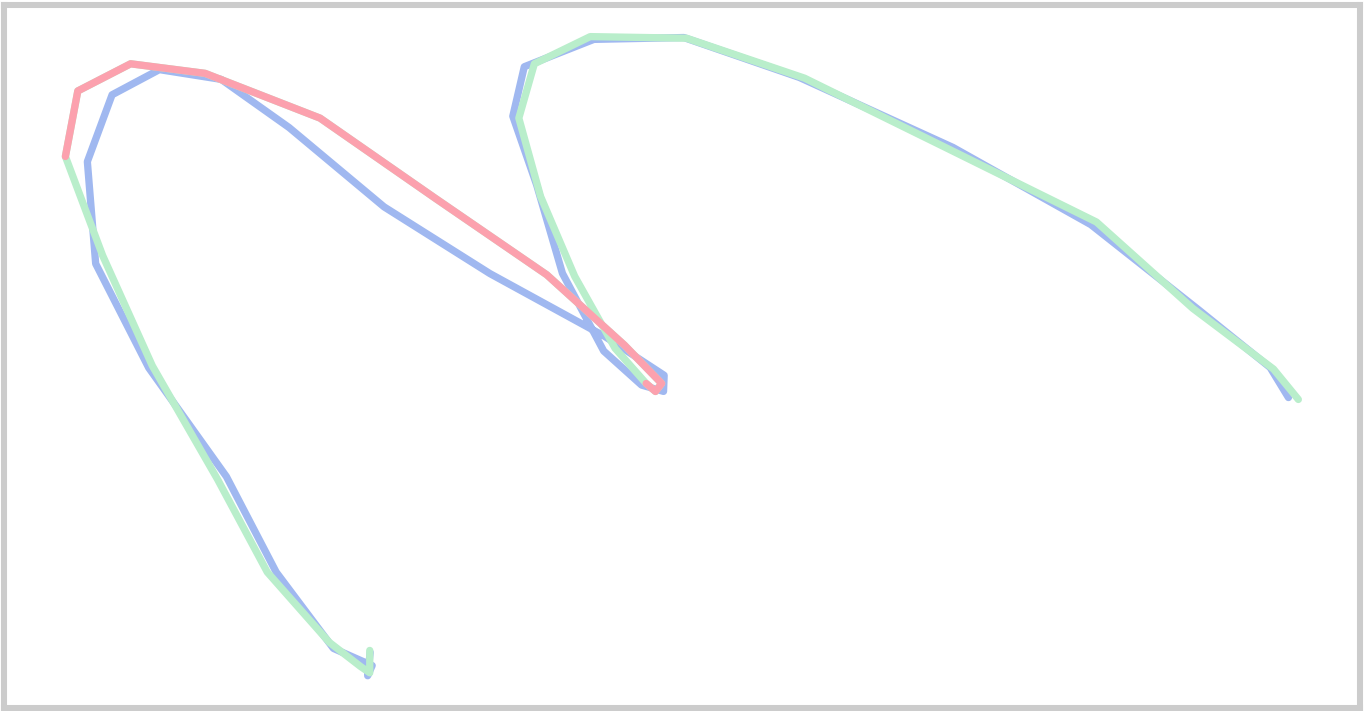} & \addpic{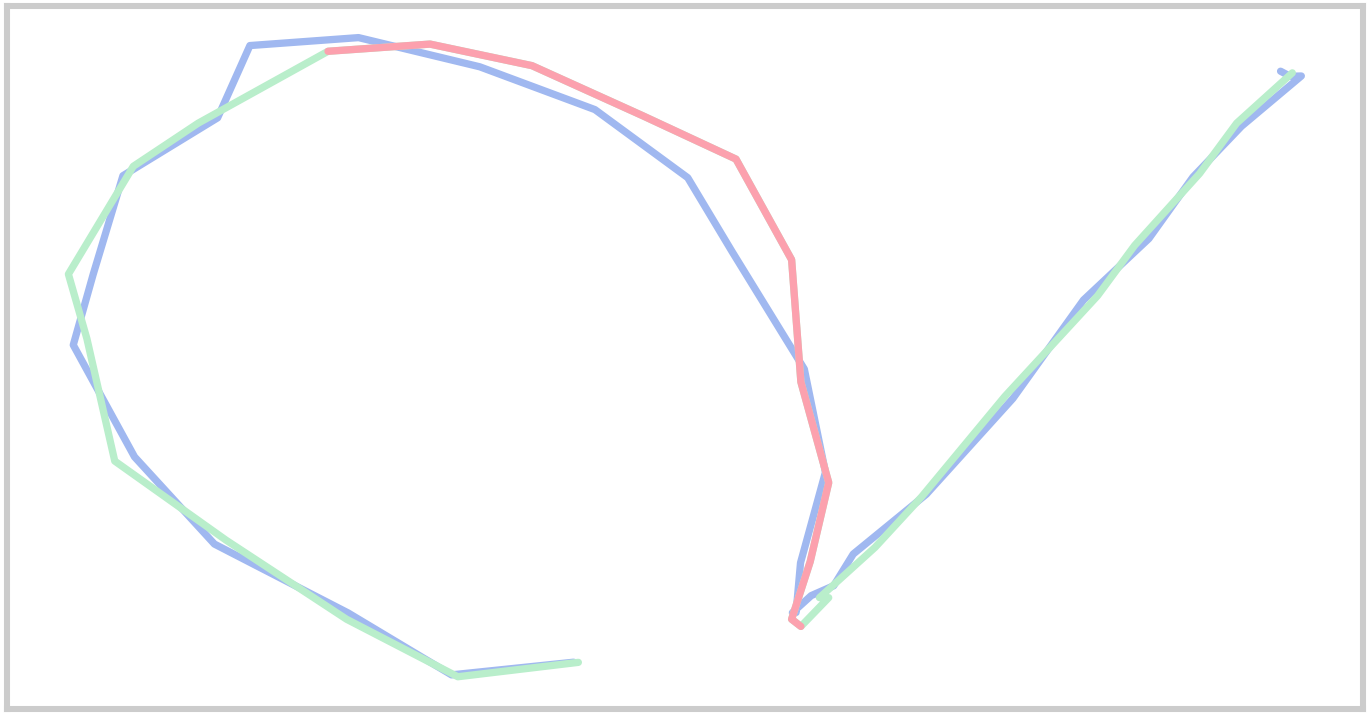} & \addpic{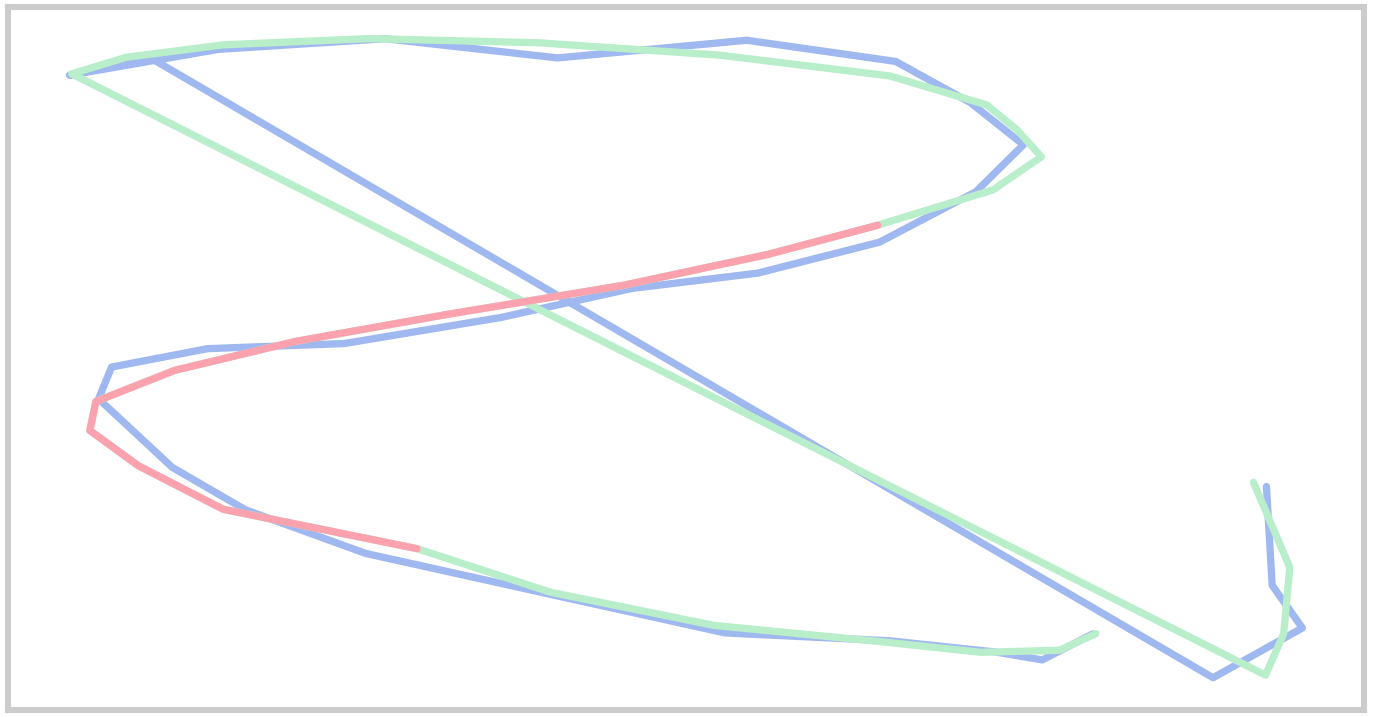} & \addpic{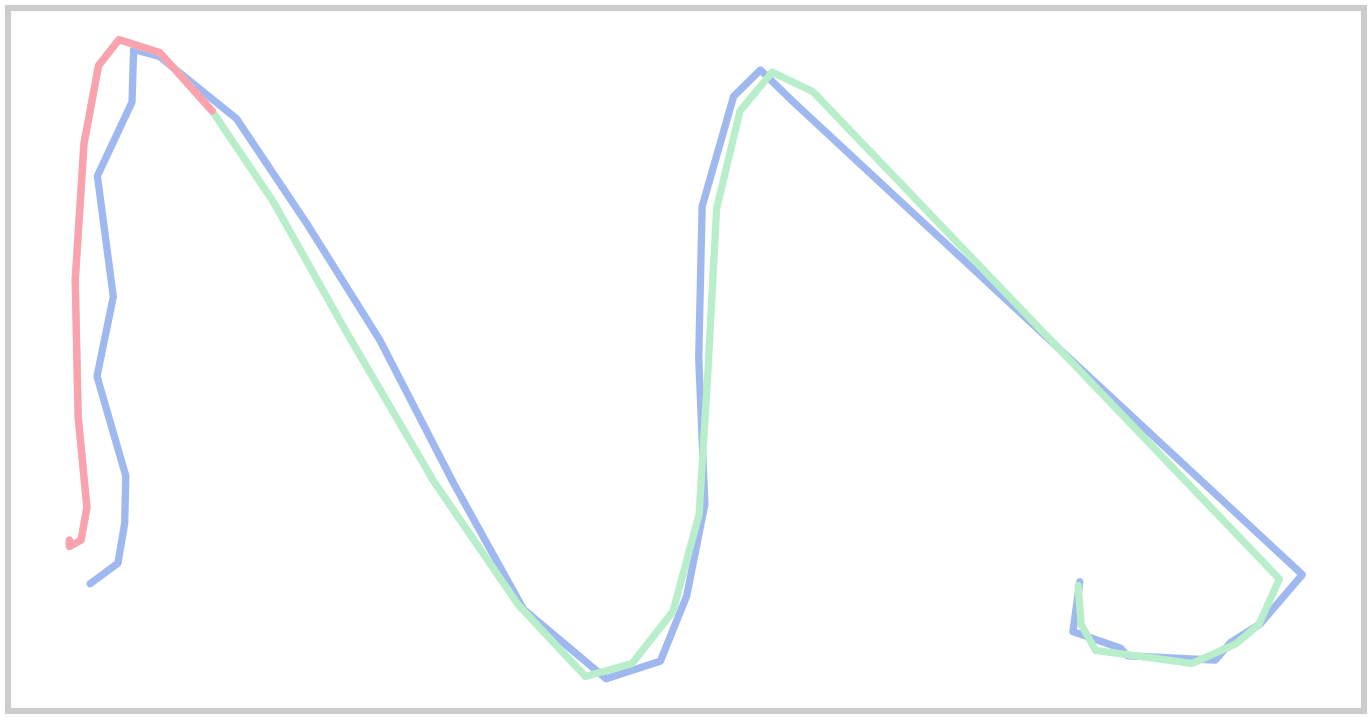}\\
        full.x.y.t & \addpic{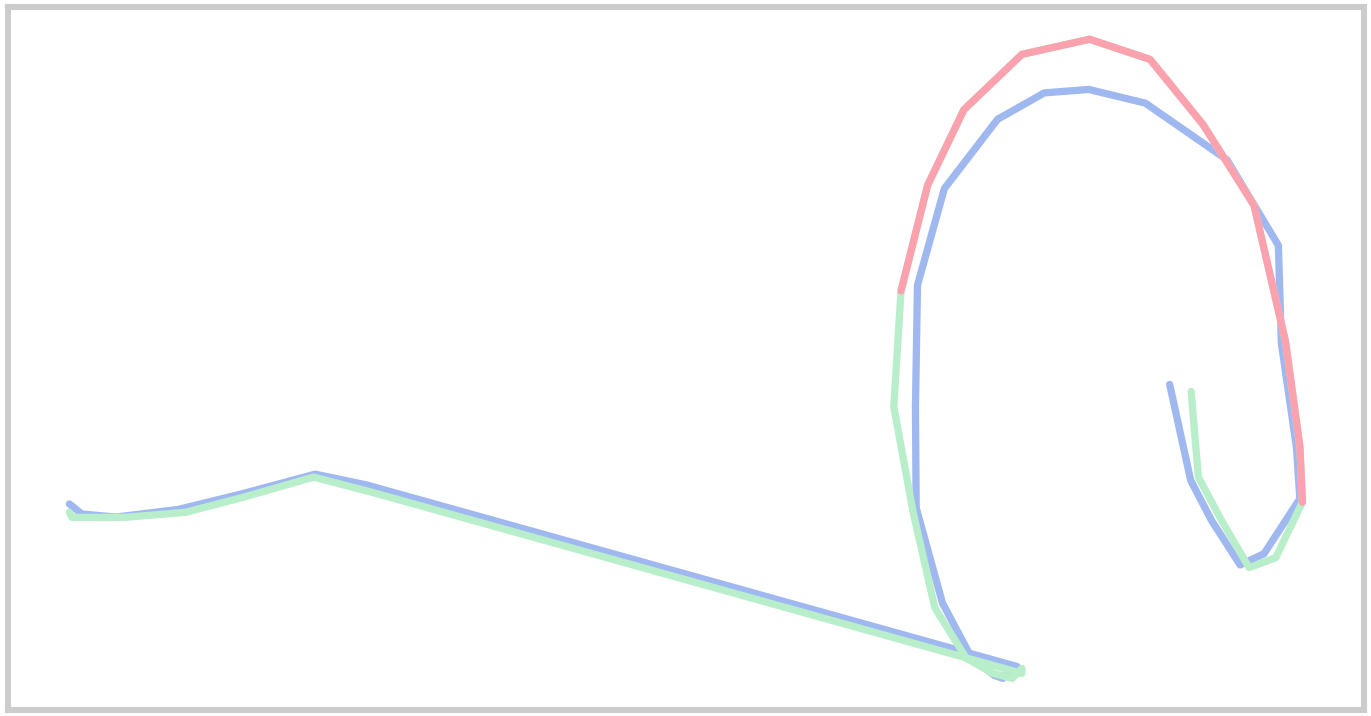} & \addpic{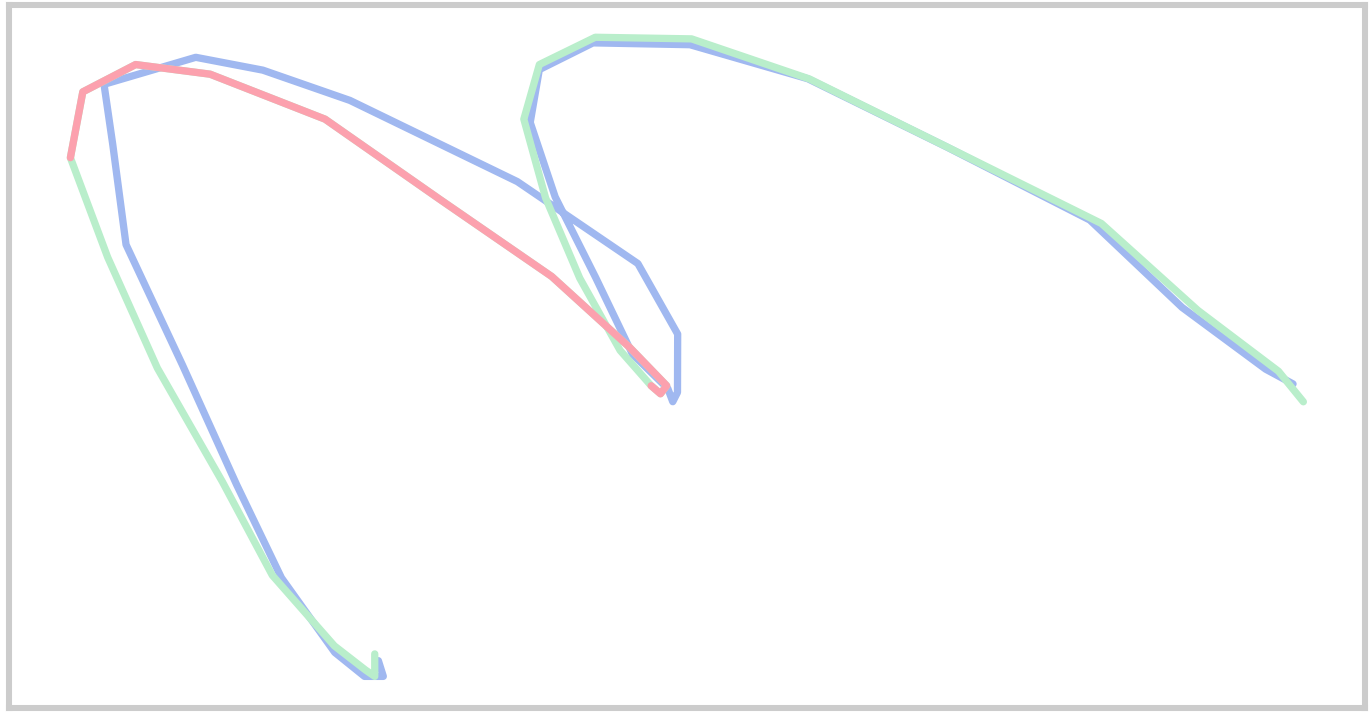} & \addpic{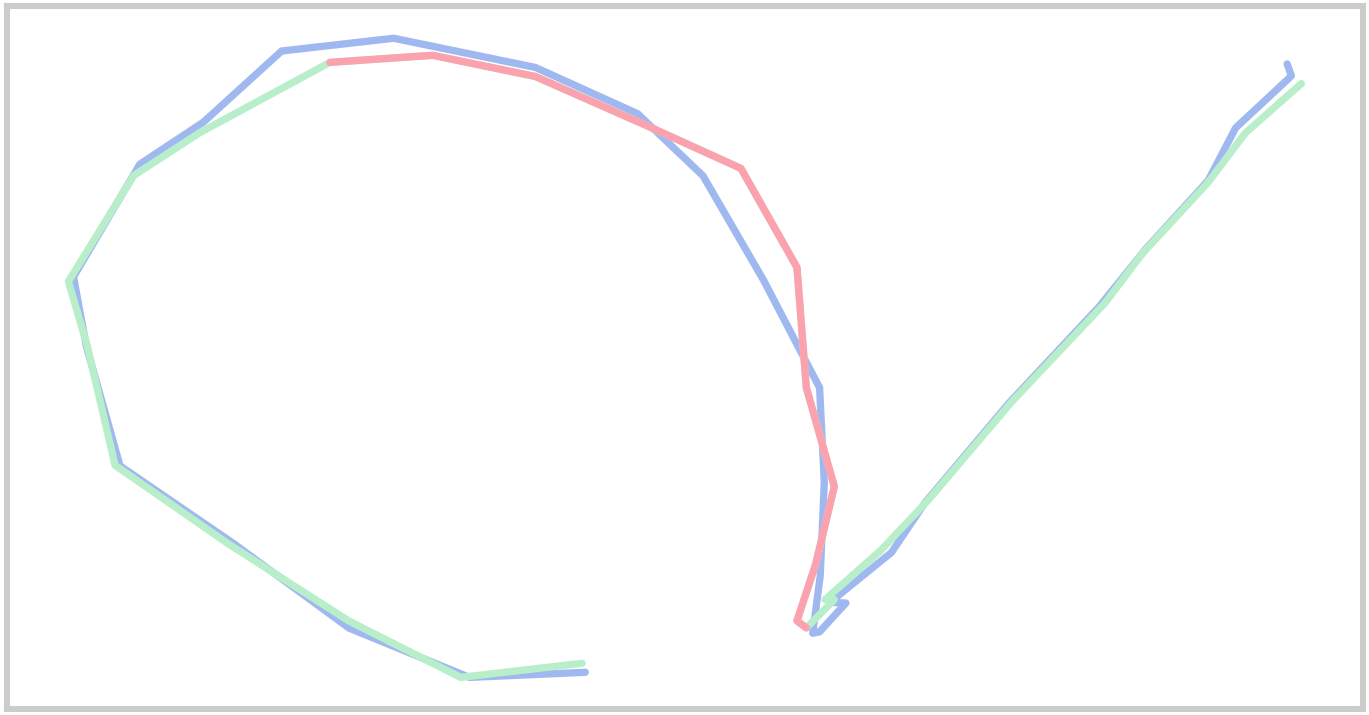} & \addpic{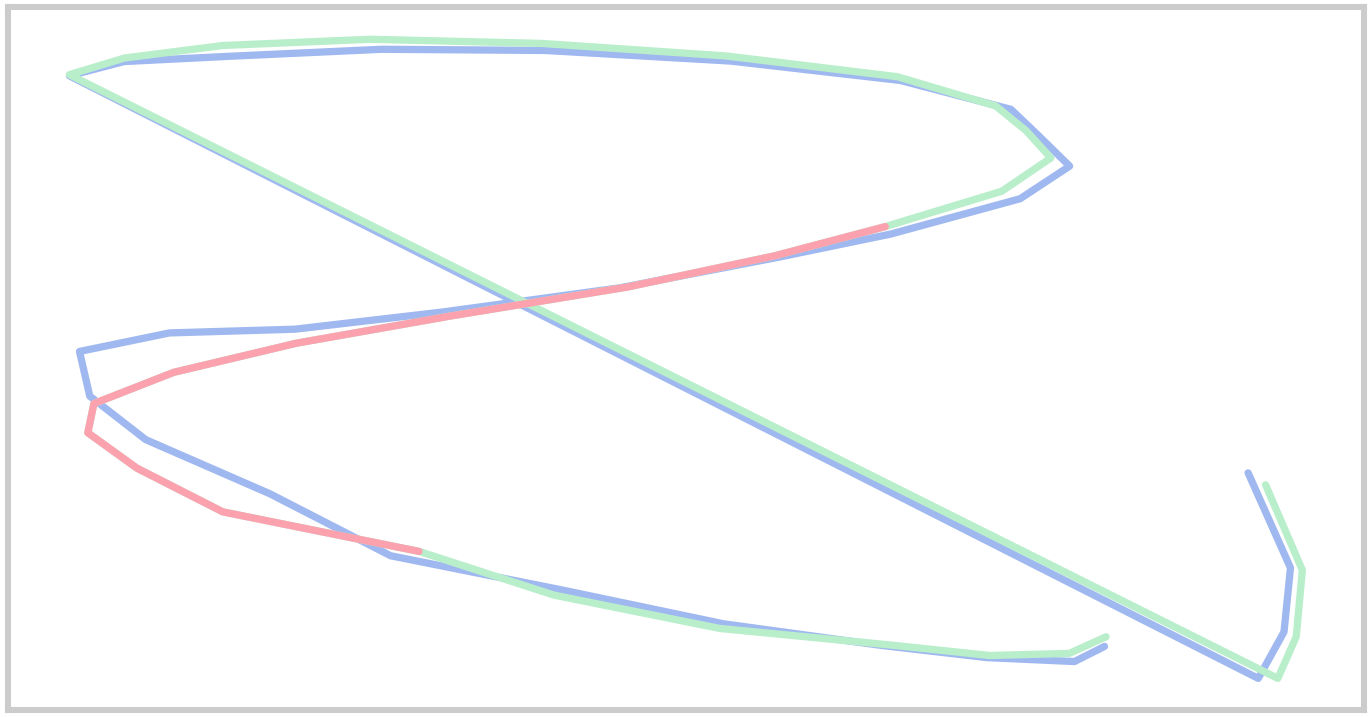} & \addpic{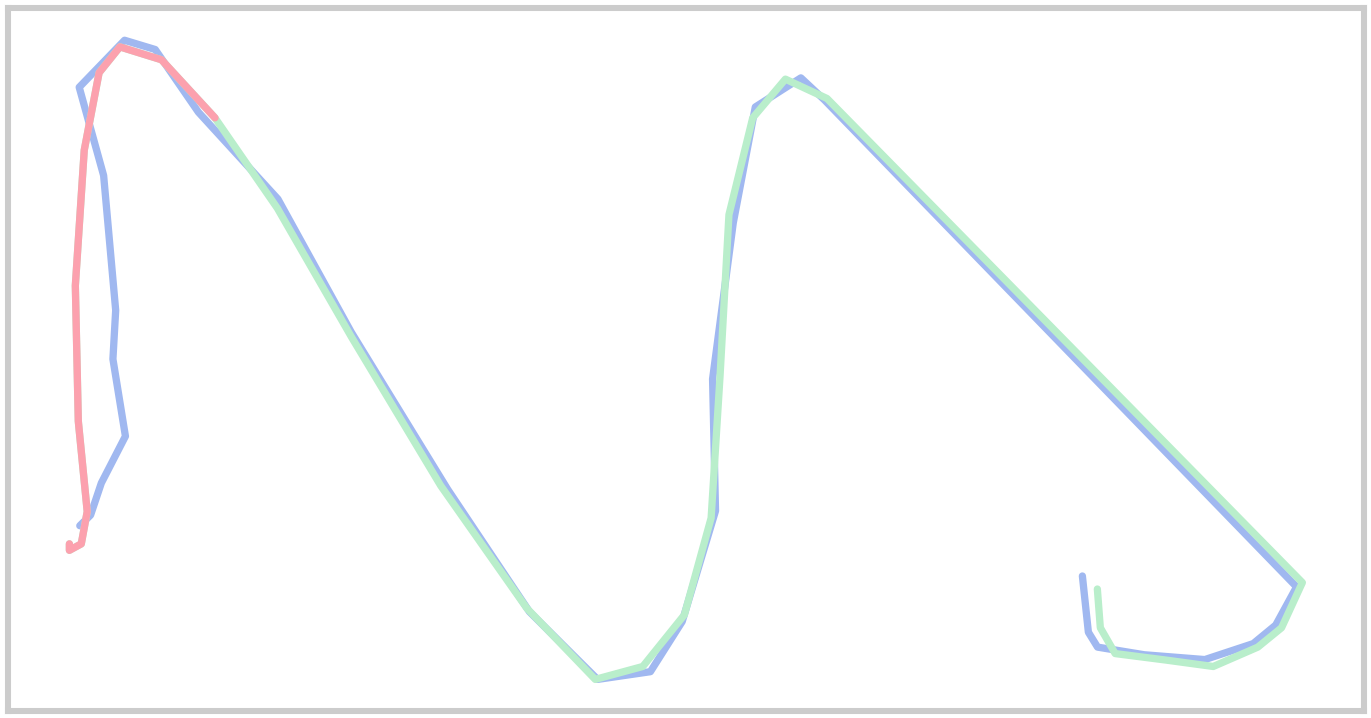}\\
        full.x.y.$\Delta t$ & \addpic{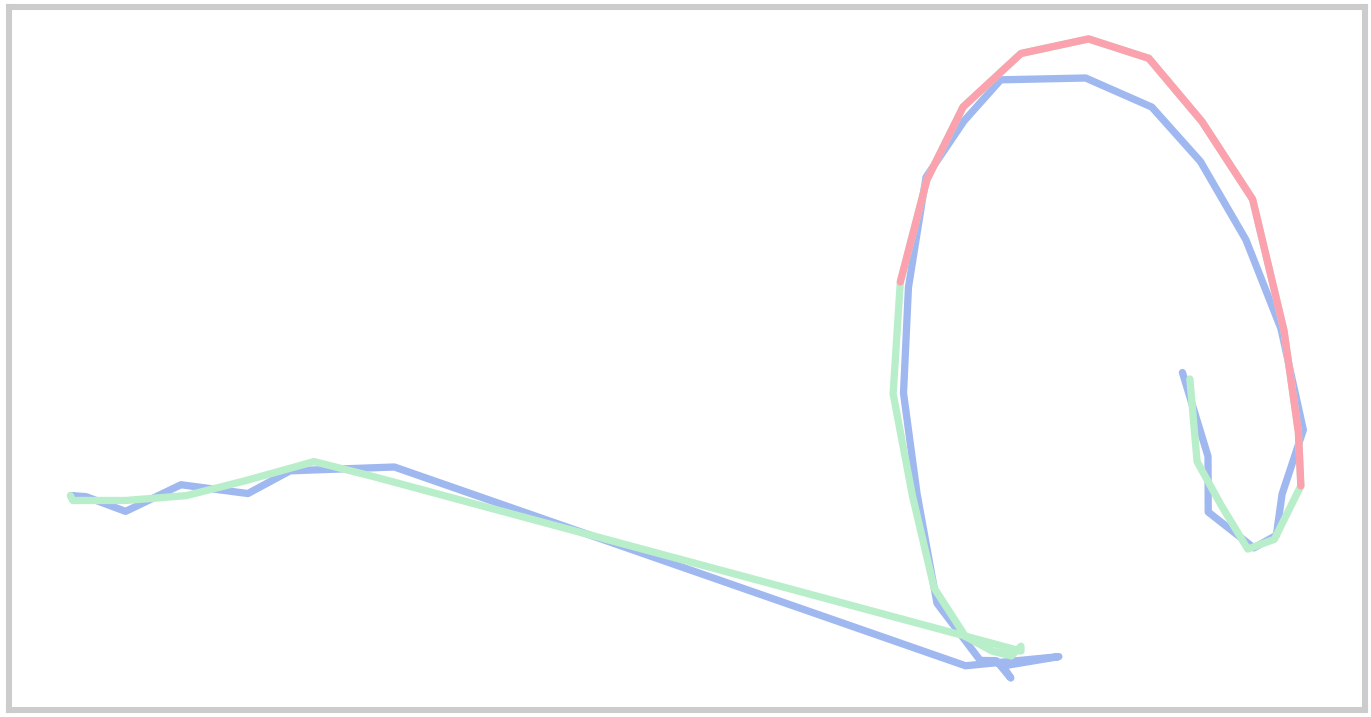} & \addpic{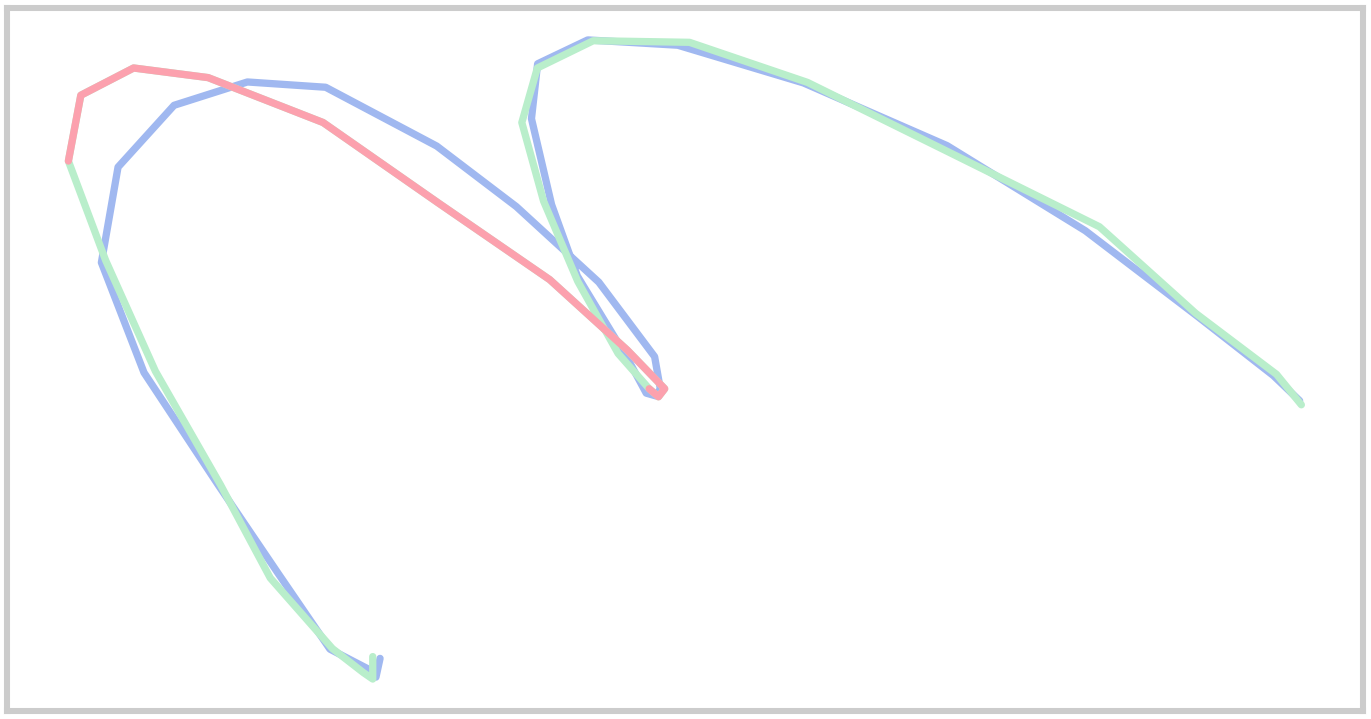} & \addpic{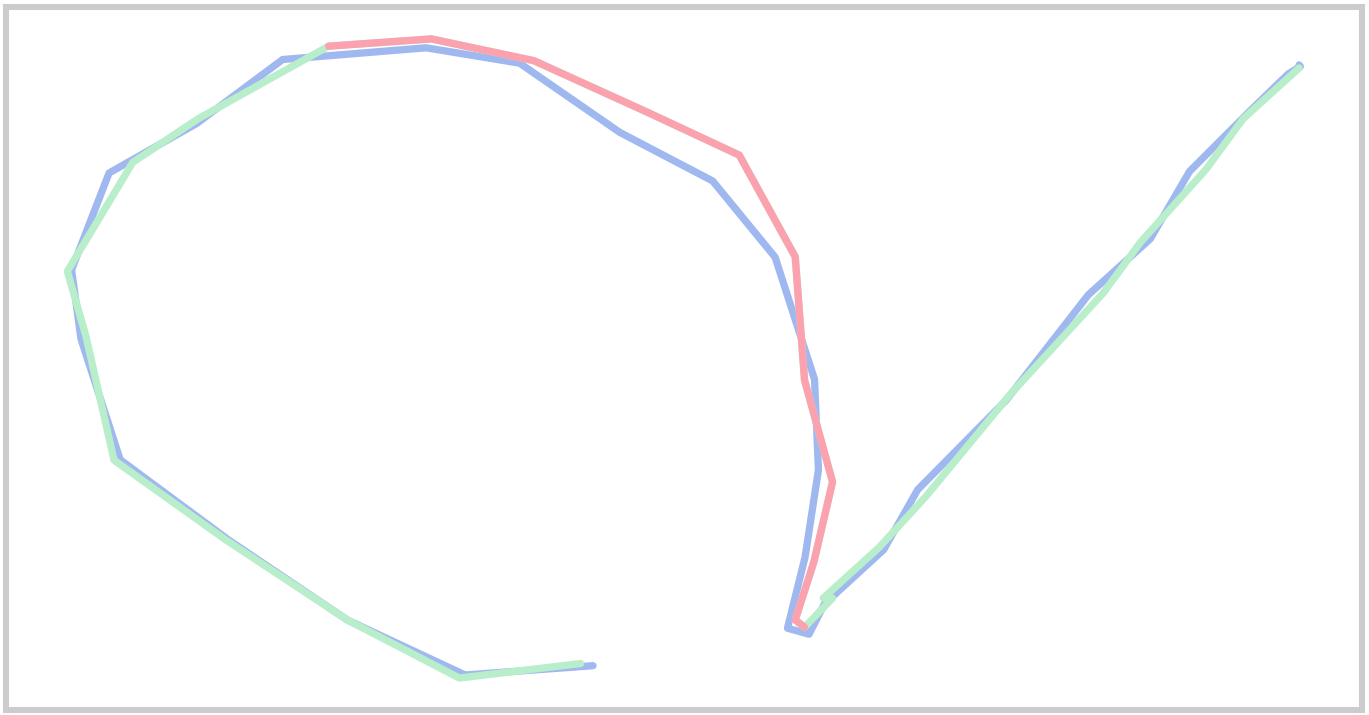} & \addpic{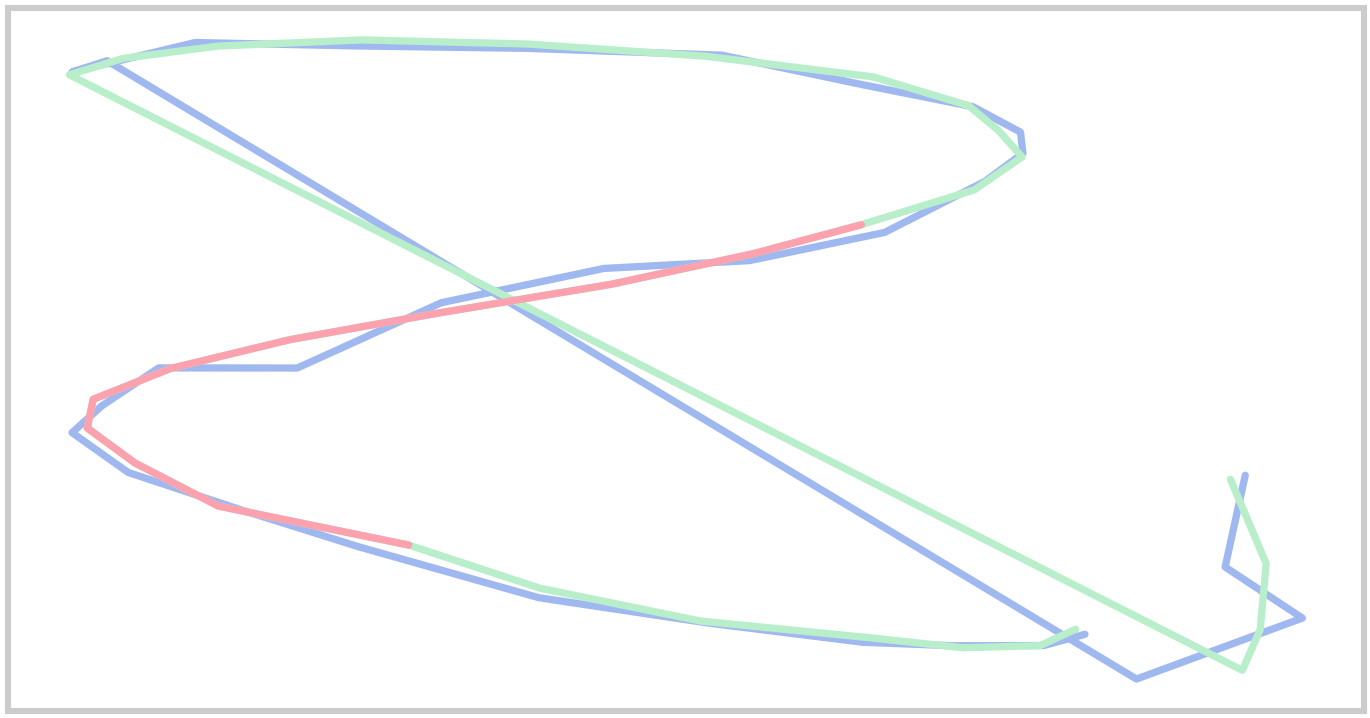} & \addpic{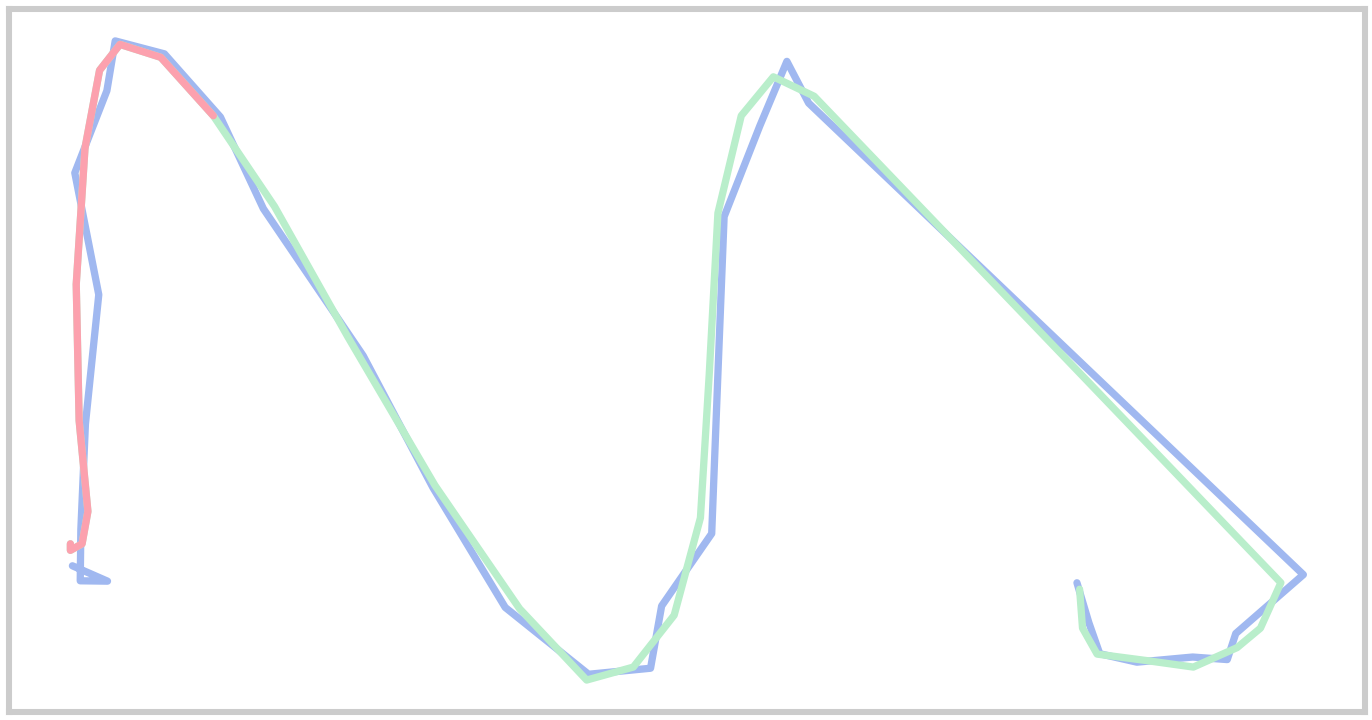}\\
        \bottomrule
    \end{tabular}
    }}
    \caption{A demonstration of the reconstructed windows using the pretrained models.}
    \label{tab:table_of_figures}
\end{figure}
%
%


\subsection{Text-Independent Writer Identification}
\label{sec:writer_id}

In this task, given a paragraph of online handwritten data, the writer should be identified. From each of the subjects in IAM-OnDB, there are eight paragraphs (strokesets) available, four of which are chosen for training. The validation and the test sets are constructed from the four paragraphs that the pretrained models have not seen in the pretraining phase (one for validation and three for test). Here, an entire paragraph should be classified. So, to utilize our pretrained models, which take a window of a fixed size as an input, each paragraph should be broken into a number of windows. To do so, we will follow the Exclusive pipeline steps to form the preprocessed train and test sets. The preprocessed validation set is constructed precisely by following the steps for the test set.  

Subsequently, the Exclusive network is fine-tuned on the preprocessed train set, and then, the resulting model is evaluated on the test set by following the steps explained in Section~\ref{sec:fine-tuning}. The reason we chose the Exclusive pipeline for the writer identification task instead of the Inclusive pipeline, which has more classification potentiality, is the computation cost. For the Inclusive network to be at its most effectiveness, at least 15 to 20 windows should be taken as input at a time. This will make the size of the model considerably larger, and it will take that much more time to train.

That said, despite the Exclusive pipeline's inferior classification potentiality compared to the Inclusive pipeline, we were still able to achieve state-of-the-art performances on this task without any manual feature engineering. These results are due to the rich representations provided by the pretrained models and the design of the Exclusive pipeline, which performs well regardless of the short length of the input sequence. 
 A more detailed description of the examined classifiers can be found in Table~\ref{tab:networks}, and the results are compared in Table~\ref{tab:writer}.

 \begin{landscape}
  \begin{table}
    \caption{Best-result classifier network structures in downstream tasks. The naming provided in this table will be used in the rest of this study for ease of reference.}
    \label{tab:networks}       
      \begin{center}
      \footnotesize{
      \begin{tabular}{r|ccccc}
      \hline\noalign{\smallskip}
      Name & Type & BLSTM Layer & \thead{Optional BLSTM\\ Block Type} & FC Layers  \\
      \noalign{\smallskip}\hline\noalign{\smallskip}
      Inc20.agg20.FC\_1 & Inclusive network & 20*tanh + dropout(0.1) & Aggregate-state & \parbox{6cm}{\centering 20*linear + BN + dropout(0.2) + 2*sigmoid}\\
      Inc20.agg20.FC\_2 & Inclusive network & 20*tanh & Aggregate-state & \parbox{6cm}{\centering 20*linear + BN + 2*sigmoid} \\
      Inc15.agg15.FC & Inclusive network & 15*tanh + dropout(0.1) & Aggregate-state & \parbox{6cm}{\centering 15*linear + dropout(0.25) + 2*sigmoid} \\
      Inc20.full20.FC & Inclusive network & 20*tanh + dropout(0.1) & Full-state & \parbox{6cm}{\centering 20*linear + BN + dropout(0.2) + 2*sigmoid} \\
      Inc20.FC & Inclusive network & - & Full-state & \parbox{6cm}{\centering Flatten + 20*relu + 20*relu + 2*sigmoid} \\
      Exc.agg32.FC & Exclusive network & 32*tanh & Aggregate-state & \parbox{6cm}{\centering 32*relu + 2*sigmoid} \\
      Exc.full32.FC & Exclusive network & 32*RelU + dropout(0.1) & Full-state & \parbox{6cm}{\centering BN + 222*relu + 222*sigmoid} \\
      Exc.full16.FC & Exclusive network & 16*RelU & Full-state & \parbox{6cm}{\centering BN + 222*relu + 222*sigmoid} \\
      \noalign{\smallskip}\hline
    \end{tabular}
    }
  \end{center}
  \end{table}
\end{landscape}

\begin{table}
  \caption{Test accuracy in writer identification task. ``n\_layers'' represents the number of chosen BLSTM layers of the pre-trained model. ``Trainable'' indicates the layer of the pre-trained model which is fine-tuned.}
  \label{tab:writer}       
  \begin{tabular}{lllll}
    \hline\noalign{\smallskip}
    PretrainedModel & Trainable & n\_layers & Classifier & Accuracy  \\
    \noalign{\smallskip}\hline\noalign{\smallskip}
    full.x.y & First & 1 & Exc.full16.FC & 97.80\% \\
    full.x.y & None & 2 & Exc.full16.FC & 98.14\% \\
    full.x.y.$\Delta t$  & Second & 2 & Exc.full32.FC & 98.48\% \\
    agg.x.y & Second & 2 & Exc.full16.FC & 99.15\% \\
    full.x.y.t  & Second & 2 & Exc.full32.FC & 99.32\% \\
    \textbf{full.x.y} & \textbf{Second} & \textbf{2} & \textbf{Exc.full16.FC} & \textbf{99.83}\% \\
   
    \noalign{\smallskip}\hline
  \end{tabular}
\end{table}

\subsection{Gender Classification}
\label{sec:gender}

To assess the efficacy of the learned representation in classifying the individuals' gender, an equal number of writers are chosen from each gender in IAM-onDB (same setup as in~\cite{liwicki2007}). This data is then split by a ratio of 2:1 for train and test sets, respectively. To feed the data into the pretrained model, they are broken into a number of equally-sized windows by following either the Exclusive or Inclusive pipeline as explained in Section~\ref{sec:fine-tuning}. Several classifiers are employed to fine-tune the pretrained models.

Although all the mentioned settings in Table~\ref{tab:gender} achieve state-of-the-art performance, the experiments indicate the superiority of the Inclusive pipeline in contrast to the Exclusive pipeline. Moreover, it is shown that a reliable setting in the case of employing an Inclusive pipeline is to choose as many as 20 windows. Despite the efficiency of the learned representations in the pretrained models, fine-tuning the final layer can enhance the results remarkably.

Given the fact that the Inclusive pipeline with a Full-state block has more information-passing capability, this result satisfies the expectations. In spite of the combination being prone to overfitting in case of inadequate data, this is not the case in gender classification and there is more amount of data in each class compared to the case of handedness classification.

\begin{table}[h]
  \caption{Test accuracy in gender classification task. ``n\_layers'' represents the number of chosen BLSTM layers of the pre-trained model. ``Trainable'' indicates the layer of the pre-trained model which is fine-tuned.}
  \label{tab:gender}       
  \begin{tabular}{lllll}
    \hline\noalign{\smallskip}
    Pretrained Model & Trainable & n\_layers & Classifier & Accuracy  \\
    \noalign{\smallskip}\hline\noalign{\smallskip}
    full.x.y.$\Delta t$ & None & 2 & Inc15.agg15.FC & 69.29\% \\
    full.x.y & None & 2 & Exc.agg32.FC & 74.45\% \\     
    agg.x.y & None & 1 & Inc20.agg20.FC\_1 & 74.79\% \\
    full.x.y & None & 1 & Inc20.agg20.FC\_1 & 75.33\% \\
    full.x.y.$\Delta t$ & None & 2 & Inc20.agg20.FC\_2 & 75.88\% \\
    \textbf{full.x.y} & \textbf{Second} & \textbf{2} & \textbf{Inc20.full20.FC } & \textbf{79.94\%} \\
    \noalign{\smallskip}\hline
  \end{tabular}
\end{table}



\subsection{Handedness Classification}
\label{sec:handedness}

Handedness classification is essentially the task of identifying whether a writer is left-handed or right-handed. In spite of inherent similarities in the tasks of gender and handedness classification, data bias is much of a challenge in the handedness classification task. Following the rarity of left-handed individuals in society, IAM-OnDB also includes as few as 20 left-handed individuals. Consequently, following the same data preparation pipeline as in~\cite{liwicki2007} is not effective, since the total amount of training data will not be adequate to train a neural network of interest. As a result, upon grouping the data based on individuals' handedness, it is then split with a ratio of 3:1 in each group to construct the train and test sets, respectively.

In a similar manner to gender classification, a number of classifiers are employed in both Exclusive and Inclusive pipelines. The results are almost identical in terms of effectiveness to the ones in gender classification. However, in contrast to the gender classification task, in which an Inclusive structure with a Full-state block achieves the highest accuracy, an Inclusive structure with an Aggregate-state block results in the best outcome (Table~\ref{tab:hand}).

\begin{table}[h]
  \caption{Test accuracy in handedness classification task. ``n\_layers'' represents the number of chosen BLSTM layers of the pre-trained model. ``Trainable'' indicates the layer of the pre-trained model which is fine-tuned.}
  \label{tab:hand}       
  \begin{tabular}{llllll}
    \hline\noalign{\smallskip}
    Pretrained Model& Trainable & n\_layers & Classifier & Accuracy & F1 \\
    \noalign{\smallskip}\hline\noalign{\smallskip}
    full.x.y & None & 2 & Exc.agg32.FC & 92.70\% & 96.13\% \\
    full.x.y & Second & 2 & Inc20.full20.FC & 93.03\% & 96.38\% \\
    agg.x.y & None & 1 & Inc20.agg20.FC\_1 & 93.81\% & 96.79\% \\
    full.x.y.$\Delta t$  & None & 2 & Inc15.agg15.FC & 93.84\% & 96.72\% \\
    \textbf{full.x.y} & \textbf{None} & \textbf{1} & \textbf{Inc20.agg20.FC\_1} & \textbf{96.13\%} & \textbf{98.00\%} \\
    \noalign{\smallskip}\hline
  \end{tabular}
\end{table}

Although the Inclusive structure with a Full-state block results in more prediction potentiality, it is prone to overfitting if there is a low amount of data. As mentioned earlier, the data in handedness classification is biased and there are a few left-handed instances. As a result, it is not expected that using a Full-state block performs the best; rather, an Aggregate-state block will reduce the network complexity and result in better classification on this low amount of data.


\subsection{The Effects Of The Pretraining on IAM-onDB}

To measure the pretrained models' contributions to achieving state-of-the-art results on the downstream tasks, we run several tests in which, instead of using the weights from the pretrained models, the weights are initialized randomly. This way, it can be examined how much the representations from the pretrained models are impactful.

For each downstream task, the network structures for the top two results are chosen. Next, these structures are trained on their corresponding downstream tasks from scratch. Table~\ref{tab:effect} demonstrates the results. In all cases, there is a meaningful margin between the models that use the weights from the pretrained models and those that do not. This depicts the effectiveness of POSM in learning high-quality representations from online handwritten data.

\begin{table}[h]
  \caption{Comparison of utilizing the pretrained models and training the same network structure from scratch.}
  \label{tab:effect}       
  \footnotesize{
  \begin{tabular}{r|cccc}
    \hline
    Task & \thead{Pretrained\\ Model} & Classifier & \thead{Accuracy\\(utilizing\\ pretrained\\ model)} & \thead{Accuracy\\(from scratch)} \\
    \noalign{\smallskip}\hline\noalign{\smallskip}
    Writer Identification & full.x.y & Exc.full16.FC & 99.83\% & 95.94\% \\
    Writer Identification & full.x.y.t & Exc.full32.FC & 99.32\% & 95.43\% \\
    \noalign{\smallskip}\hline\noalign{\smallskip}
    Gender Classification & agg.x.y & Inc20.agg20.FC\_1 & 74.79\% & 69.58\% \\
    Gender Classification & full.x.y & Inc20.full20.FC & 79.94\% & 70.65\% \\
    \noalign{\smallskip}\hline
    Handedness Classification & full.x.y.$\Delta t$ & Inc15.agg15.FC & 93.84\% & 90.34\% \\
    Handedness Classification & full.x.y & Inc20.agg20.FC\_1 & 96.13\% & 90.78\% \\
    \noalign{\smallskip}\hline
  \end{tabular}
   }
\end{table}


\subsection{Comparison With Existing Studies on IAM-onDB}

The results obtained in our work on the tasks of writer identification and gender classification are comparable to the results from other studies on IAM-onDB given the analogous experiment setting. In the task of handedness classification, however, the results are not easily comparable to similar studies, as the amount of the training data differs massively. The reported F1 scores in Table~\ref{tab:hand} offer more reliable evaluation criteria for the purpose of this particular task. It also facilitates the comparison of outcomes with those of other studies. 

To provide a better insight into how better the proposed methods work in all tasks compared to similar studies, Table~\ref{tab:compare} offers a distinctive overview of different methods and the amount of training data used by them. Note that in all the mentioned studies, manual feature extraction is performed, while the proposed method in this study employs no manually extracted features.

\begin{landscape}
  \enlargethispage*{20cm}
  \begin{table}[!htbp]
    \caption{Comparison of the results in different tasks and studies. The NMEF column represents the number of manually extracted features.}
    \label{tab:compare}       
 
    \footnotesize{
      \begin{tabular}{r|lllll}
      \hline\noalign{\smallskip}
      Task & Study & Domain & NMEF & Classifier & Accuracy  \\
      \noalign{\smallskip}\hline\hline\noalign{\smallskip}
      Writer Identification & \cite{schlapbach2008} & online & 3 & GMM & 88.56\% \\
      Writer Identification & \cite{WU2017} & online & 3 & HMM-M2-V1 & 95.69\% \\
      Writer Identification & \cite{schlapbach2008} & online+offline & 27 & GMM-UBM & 98.56\% \\
      Writer Identification & \cite{venugopal2018} & online & 5 & Modified Codebook descriptor & 98.82\% \\
      Writer Identification & \cite{venugopal2018sparse} & online & NEM\footnotemark & Sparse coding based descriptors \& SVM & 99.45\% \\
      Writer Identification & \cite{venugopal2019} & online & 7 & GMM \& ensemble of SVMs & 99.30\% \\
      Writer Identification & \cite{Venugopal2020} & online & NEM & Adaptive sparse representation \& ensemble of SVMs & 99.69\% \\
      Writer Identification & \cite{venugopal2021} & online & NEM & Sparse representation, pooling sparse codes, \& SVM & 99.28\% \\
      Writer Identification & \textbf{This study} & \textbf{online} & \textbf{0} & \textbf{agg.x.y|Exc.full16.FC} & \textbf{99.15\%} \\
      Writer Identification & \textbf{This study} & \textbf{online} & \textbf{0} & \textbf{full.x.y.t|Exc.full32.FC} & \textbf{99.32\%} \\
      Writer Identification & \textbf{This study} & \textbf{online} & \textbf{0} & \textbf{full.x.y|Exc.full16.FC} & \textbf{99.83\%} \\
      \noalign{\smallskip}\hline
      Gender Classification & \cite{liwicki2007} & online & 27 & GMM & 67.06\% \\
      Gender Classification & \cite{liwicki2011} & online+offline & 36 & GMM & 67.57\% \\
      Gender Classification & \textbf{This study} & \textbf{online} & \textbf{0} & \textbf{full.x.y.$\Delta t$|Inc15.agg15.FC} & \textbf{69.29\%} \\
      Gender Classification & \textbf{This study} & \textbf{online} & \textbf{0} & \textbf{agg.x.y|Inc20.agg20.FC} & \textbf{76.69\%} \\
      Gender Classification & \textbf{This study} & \textbf{online} & \textbf{0} & \textbf{full.x.y|Inc20.full20.FC\_1} & \textbf{79.94\%} \\
      \noalign{\smallskip}\hline
      Handedness Classification & \cite{liwicki2007} & online & 27 & GMM & 84.66\% \\
      Handedness Classification & \textbf{This study} & \textbf{online} & \textbf{0} & \textbf{full.x.y|Inc20.agg20.FC\_1} & \textbf{96.13\%}\footnotemark \\
      \noalign{\smallskip}\hline
      
    \end{tabular}
      }
  \end{table}
  \footnotetext[1]{NEM = Not Explicitly Mentioned}
  \footnotetext{Given the difference in experiment setups of the two studies, also look at F1-scores reported in Table~\ref{tab:hand} for a better comparison.}
\end{landscape}


\subsection{Experiments on the CASIA online handwriting dataset}
\label{sec:Experiments_on_CASIA}

In this section, we will conduct experiments to demonstrate the effectiveness of the proposed methods in producing meaningful representations by showing the results of the English and Chinese text-independent writer identification tasks when the weights of the model pretrained on the POSM pretext task are utilized compared to training from the scratch. Subsequently, the results are compared to other state-of-the-art approaches. Since our aim is to validate the efficacy of the representations learned with the POSM pretext task on online handwritten text, we have only considered the $xy$-coordinates of the pen-down movements (where the pressure is greater than zero).


\begin{table}[h]
  \caption{Test accuracy on the writer identification task of the English dataset of the CASIA dataset. ``n\_layers'' represents the number of chosen BLSTM layers of the pre-trained model. ``Trainable'' indicates the layer of the pre-trained model which is fine-tuned.}
  \label{tab:CASIA_English}       
  \begin{tabular}{lllll}
    \hline\noalign{\smallskip}
    \thead{Pretrained Model/\\ from scratch} & Trainable & n\_layers & Classifier & Accuracy  \\
    \noalign{\smallskip}\hline\noalign{\smallskip}
    from scratch & ------ & ------ & Exc.full16.FC & 93.28\% \\
    \textbf{full.x.y} & \textbf{Second} & \textbf{2} & \textbf{Exc.full16.FC } & \textbf{99.25\%} \\
    \noalign{\smallskip}\hline
  \end{tabular}
\end{table}

The experimental setting for the English dataset is the same as \cite{zhang2017}, where the two free-content pages from each writer are used for training and the fixed-content pages are used for testing. Given that the full.x.y pretrained model along with the Exc.full16.FC classifier led to the best result on the writer identification task on the IAM-onDB dataset, the same pairing is utilized to train a model on the English dataset. The preprocessed train and test sets are constructed and fine-tuned as explained in Section~\ref{sec:fine-tuning}. Then, as a measure of comparison, a model with the same structure is trained from the scratch, without using the weights of the pretrained model. The results are shown in Table~\ref{tab:CASIA_English}. This means that the representations learned by using the POSM pretext task on the IAM-onDB dataset are impactful when utilized on the English dataset of the CASIA.

\begin{table}[h]
  \caption{Test accuracy on the writer identification task of the Chinese dataset of the CASIA dataset. ``n\_layers'' represents the number of chosen BLSTM layers of the pre-trained model. ``Trainable'' indicates the layer of the pre-trained model which is fine-tuned.}
  \label{tab:CASIA_Chinese}       
  \begin{tabular}{lllll}
    \hline\noalign{\smallskip}
    \thead{Pretrained Model/\\ from scratch} & Trainable & n\_layers & Classifier & Accuracy  \\
    \noalign{\smallskip}\hline\noalign{\smallskip}
    from scratch & ------ & ------ & Exc.full16.FC & 92.54\% \\
    \textbf{full.x.y} & \textbf{Second} & \textbf{2} & \textbf{Exc.full16.FC } & \textbf{99.47\%} \\
    \noalign{\smallskip}\hline
  \end{tabular}
\end{table}

As for the Chinese dataset, the same experimental setting as the English dataset is used where the preprocessed train set is constructed from the two free-content pages, and the preprocessed test set is constructed from the fixed-content pages. However, when using the weights from the full.x.y pretrained model, which is learned on the IAM-onDB dataset, no considerable improvements were offered compared to the model with the same structure trained from the scratch. This can be explained by the fact that the pretrained model is trained on the English language and then the model is being asked to provide suitable representations of the Chinese language which has drastically different writing styles and patterns. To display that the POSM pretext task can be used to learn meaningful representations from the Chinese language as well, a model with the same structure and set of features as the full.x.y is pretrained on the two free-content pages of the Chinese dataset by following the steps explained in the Section~\ref{sec:pretraining}. This model is named Chinese.full.x.y. Next, the Chinese.full.x.y along with the Exc.full16.FC classifier is trained on the preprocessed train set. Subsequently, as a measure of comparison, a model with the same structure is trained from the scratch without using the weights of the pretrained model. The results on the preprocessed test set are presented in Table~\ref{tab:CASIA_Chinese}. The improvements provided by the pretrained model signify the competency of the POSM pretext task and the proposed methods to learn meaningful representations regardless of the language that is used for the pretraining phase. Table~\ref{tab:CASIA_Comparison} demonstrates a comparison of the results with other studies. Note that in this study, only the $xy$-coordinate of the pen-down movements have been used while no hand-crafted features have been extracted.

\begin{table}[h]
  \caption{Comparison of the results of the text-independent writer identification task, on the  English and Chinese datasets of the CASIA online handwriting dataset.}
  \label{tab:CASIA_Comparison}       
  \begin{tabular}{lll}
    \hline\noalign{\smallskip}
    Study & English dataset & Chinese dataset \\
    \noalign{\smallskip}\hline\noalign{\smallskip}
    Liwicki et al. ~\cite{Liwicki_CASIA} & 82.00\% & 80.00\% \\
    Bulacu et al. ~\cite{Bulacu_CASIA} & 85.00\% & 84.00\% \\
    Li et al. ~\cite{li2009} & 93.60\% & 91.50\% \\
    Venugopal et al. ~\cite{venugopal2018} & 95.68\% & 91.42\% \\
    Yang et al. ~\cite{Yang_CASIA} & 98.51\% & 95.72\% \\
    BabaAli. ~\cite{BabaAli2021} & 98.68\% & 96.03\% \\
    Zhang et al. ~\cite{zhang2017} & 100\% & 99.46\% \\
    \textbf{This study} & \textbf{99.25\%} & \textbf{99.47\%} \\
    \noalign{\smallskip}\hline
  \end{tabular}
\end{table}


 


\section{Discussion}
\label{sec:discussion}

In this research, we offered a self-supervised representation learning perspective in analyzing online handwritten text, evaluated mainly on classification downstream tasks. For the specific case of online handwritten data, extracting rich representations can be obtained by defining appropriate pretext tasks according to pen location and timing data. Part of Stroke Masking (POSM) is the one that has been introduced in this study, but it is not the only possible approach.

We believe that the methods proposed in this study will work quite well on most of the classification tasks on the online handwritten text modality.
This is due to the reason that the overall pattern of many classification tasks and datasets is that the number of subjects is limited (less than 200 and in many cases less than 100) and the data available from each subject is limited to a number of paragraphs or a handful of predefined tasks completed by the subject such as writing some words or sentences in the specified boxes.
At the same time, the captured sequences corresponding to each subject are quite long. This is due to the high sampling rates of the capturing devices which could result in some sequences of length above 10,000 for just one paragraph written by a subject (or in the case of the predefined tasks, a single task completed by a subject). 

The long length of each sequence combined with the few number of subjects raises an issue since the RNN-based approaches will suffer from the vanishing gradient problem, and also it is not easily feasible to train transformer-based networks from scratch with a small number of subjects and a limited amount of data from each of them. Due to this obstacle, many have based their approach mainly on manually extracting features from the sequences and utilizing the extracted features to train a model and then classify. These approaches suffer from having to manually define and extract features, which might not be as optimal as a rich representation learned by a pretrained model.

The pretrained model and the fine-tuning methods proposed in this study will aid in overcoming these shortcomings by not depending on manually extracted features and instead, using the output representations of the pretrained model. Also, by breaking each long sequence into a number of windows, the vanishing gradient problem is avoided. However, by breaking the sequence into smaller chunks, the context provided for the model in a single window will be limited. To mitigate this issue, the Inclusive pipeline has been proposed which takes in an arbitrary number of windows ($n_{windows}$) with no overlap, at the same time. As the  $n_{windows}$ parameter increases, the computation cost will become greater, but in turn it is possible to increase the shifting parameter in the fine-tuning process ($s_{train}$ for the training phase and $s_{test}$ for the test phase) which decreases the overall computation cost. Finding the best balance between the trade-off of performance and computation cost might be different corresponding to the task at hand.

The experiments conducted in this study are mainly focused on classification tasks. However, a more complicated task could be handwriting recognition in which the model should recognize the sequence of characters written by the individuals. The question is can POSM benefit this downstream task as well? To answer this question, let's revisit the reconstruction task. This intrinsic downstream evaluation criterion depicts the ability of the pretrained model to reconstruct the masked parts of unseen data. Empirical findings suggest that the model is able to reconstruct the masked part corresponding to a character with good precision. As a result, the model might also be effective in a higher-level task of handwriting recognition. However, to fully address this potential, a future work would be dedicated to this assessment.

\section{Conclusion}
\label{sec:conclusion}

Designing a pretext task for the purpose of self-supervised representation learning is challenging, domain-dependent, and modality-specific. Consequently, learning rich representations to solve online handwriting-related tasks requires the proposal of a suitable pretext task. Part of Stroke Masking, or POSM for short, is the first pretext task introduced for the domain of online handwritten data. The pretrained models trained on POSM provide high-quality representations by masking blocks of a sequence of data points and trying to reconstruct the missing parts. By suggesting the appropriate pipelines and structures for fine-tuning the pretrained models on various downstream tasks, these learned representations are shown to be impactful.
The downstream tasks studied in this body of work are text-independent writer identification, gender classification, and handedness classification. State-of-the-art results are obtained in all the mentioned tasks performed on IAM-OnDB and the representations learned from the POSM pretext task are shown to be impactful on both the Chinese and English datasets of the BIT CASIA database.


\bibliography{references}

\end{document}